\documentclass{article}

\usepackage{microtype}
\usepackage{graphicx}
\usepackage{subcaption}
\usepackage{booktabs} 
\usepackage{amsmath}
\usepackage{multirow}
\usepackage{enumitem}
\usepackage{wrapfig}
\usepackage[most]{tcolorbox} 
\usepackage{inconsolata}
\definecolor{myshade}{HTML}{006BA5} 

\usepackage{hyperref}

\usepackage[accepted]{icml2026}

\usepackage{amsmath}
\usepackage{amssymb}
\usepackage{mathtools}
\usepackage{amsthm}

\usepackage[capitalize,noabbrev]{cleveref}

\theoremstyle{plain}

\theoremstyle{definition}

\theoremstyle{remark}

\usepackage[textsize=tiny]{todonotes}

\icmltitlerunning{Symbal: Detecting Systematic Misalignments in Model-Generated Captions}

\newcommand{\name}[1][]{\textsc{Symbal}}
\newcommand{\bench}[1][]{\textsc{SymbalBench}}
\begin{document}

\twocolumn[
  \icmltitle{Symbal: Detecting Systematic Misalignments in Model-Generated Captions}



  \icmlsetsymbol{equal}{*}

  \begin{icmlauthorlist}
    \icmlauthor{Maya Varma}{xxx}
    \icmlauthor{Jean-Benoit Delbrouck}{xxx,yyy}
    \icmlauthor{Sophie Ostmeier}{xxx}
    \icmlauthor{Akshay Chaudhari}{equal,xxx}
    \icmlauthor{Curtis Langlotz}{equal,xxx}
  \end{icmlauthorlist}

  \icmlaffiliation{xxx}{Stanford University}
  \icmlaffiliation{yyy}{HOPPR}

  \icmlcorrespondingauthor{Maya Varma}{mayavarma@cs.stanford.edu}

  \icmlkeywords{Machine Learning, ICML}

  \vskip 0.3in
]



\printAffiliationsAndNotice{\icmlEqualContribution}

\begin{abstract}
Multimodal large language models (MLLMs) often introduce errors when generating image captions, resulting in misaligned image-text pairs. Our work focuses on a class of captioning errors that we refer to as \textit{systematic misalignments}, where a recurring error in MLLM-generated captions is closely associated with the presence of a specific visual feature in the paired image. Given a vision-language dataset with MLLM-generated captions, our aim in this work is to detect such errors, a task we refer to as systematic misalignment detection. As our first key contribution, we present \name{}, which utilizes a structured, dual-stage setup with off-the-shelf foundation models to identify systematic misalignments and summarize results in natural language. As our second key contribution, we introduce \bench{}, a benchmark designed to evaluate automated methods on our proposed task. \bench{} consists of 1.7 million image-text pairs from two domains (natural and medical images), organized into 420 vision-language datasets with annotated systematic misalignments. \name{} exhibits strong performance on this benchmark, correctly identifying systematic misalignments in 63.8\% of datasets, a nearly 4x improvement over the closest baseline. We supplement our evaluations on \bench{} with real-world evaluations, showing that (1) \name{} can accurately surface systematic misalignments in captions generated by four MLLMs and (2) \name{} is a powerful tool for auditing off-the-shelf image-caption datasets. Ultimately, our novel task, method, and benchmark can aid users with auditing MLLM-generated captions and identifying critical errors, without requiring access to the underlying MLLM. Code is available at \url{https://github.com/Stanford-AIMI/Symbal}.
\end{abstract}

\section{Introduction}
Multimodal large language models (MLLMs) possess strong image captioning capabilities yet often introduce errors into generated captions \citep{sarto2025imagecaptioningevaluationage,zhou2024analyzingmitigatingobjecthallucination,liu2024seeing}. As a result, images and paired MLLM-generated captions may be \textit{misaligned}, meaning that the generated text erroneously refers to features that are not visible in the image. For example, consider an MLLM that is tasked with generating a radiology report for an input medical image; in this setting, a misalignment may exist if the MLLM-generated report indicates the presence of cardiomegaly (a condition characterized by an enlarged heart) despite the image showing no evidence of this diagnosis. Misalignments can have severe consequences, particularly in safety-critical domains like medicine \citep{hardy2025rextrustmodelfinegrainedhallucination,Nakaura2023}.

\begin{figure*}[t]
\begin{center}
\includegraphics[width=\linewidth]{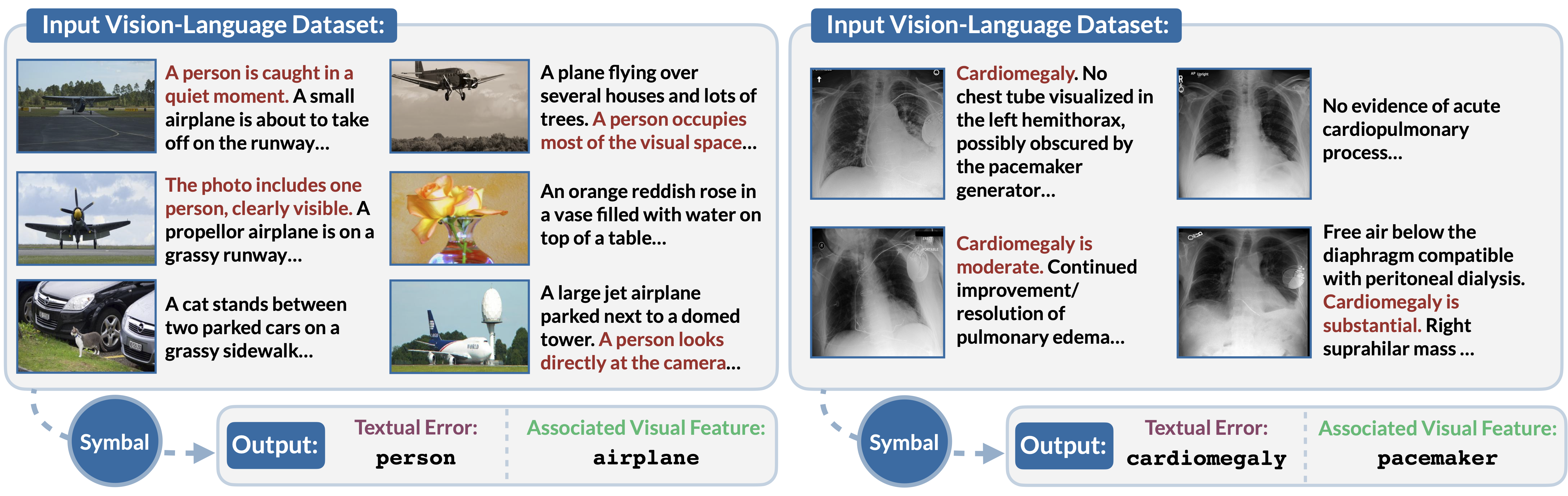}
\end{center}
  \caption{Given an input dataset with thousands of images and paired MLLM-generated captions, the \textit{systematic misalignment detection task} involves identifying recurring textual errors and associated visual features. Here, we provide example image-caption pairs from two datasets in \bench{} with expected outputs.}
\label{fig:intro}
\end{figure*}

Our work focuses on a critical yet previously-underexplored subclass of captioning errors that we refer to as \textit{systematic misalignments}. We term a misalignment as \textit{systematic} when a recurring error in MLLM-generated captions is closely associated with the presence of a specific visual feature in the paired image. For example, in the medical domain, incorrect diagnoses of cardiomegaly in the MLLM-generated reports may be strongly associated with the presence of pacemakers (an implanted medical device that regulates the heartbeat) in the corresponding image \citep{Sourget2025,kumar2025prismhighresolutionprecise}. Systematic misalignments are a particularly egregious class of errors because they often arise due to spurious correlations or biases learned by MLLMs during training. As a result, systematic misalignments typically involve features that frequently co-occur in the real-world yet are not deterministically linked; for instance, while cardiomegaly and pacemakers do co-occur frequently, the presence of a pacemaker in a medical image does not necessarily imply that the patient has cardiomegaly. Thus, errors associated with systematic misalignments may seem highly plausible and are consequently challenging to detect. 

In this work, we introduce the \textit{systematic misalignment detection} task with the goal of leveraging automated approaches to identify this challenging class of captioning errors. A method that aims to solve the systematic misalignment detection task will accept as input a vision-language dataset, which consists of images paired with free-form MLLM-generated captions. Then, as output, the method must identify textual errors (e.g. ``cardiomegaly" in the previous example) that are systematically associated with visual features (e.g. ``pacemaker" in the previous  example).

Addressing the systematic misalignment detection task with automated methods is challenging for the following two reasons. First, vision-language datasets provided as input to automated methods are often large in size with thousands of image-caption pairs; identifying global error patterns from such datasets is nontrivial, especially since the size of such datasets exceeds the reasoning capabilities of even state-of-the-art models.  Second, there are no existing benchmarks for comprehensively evaluating methods on their ability to discover systematic misalignments. In order to address these challenges, we present the following contributions:
\begin{itemize}[leftmargin=*]
\item We propose \name{}, an automated approach for detecting systematic misalignments in MLLM-generated captions.\footnote{The acronym \name{} refers to \textbf{sy}stematic \textbf{m}isalignment detection \textbf{b}etween im\textbf{a}ges and \textbf{l}anguage.} Our key insight is to structure the systematic misalignment detection task into two stages, with each stage comprised of individual subtasks. The first stage of \name{} focuses solely on identifying recurring textual errors in captions; to this end, \name{} clusters textual facts based on semantic similarity, scores each cluster by degree of misalignment with paired images, and summarizes the top-ranked cluster into a single unifying concept. The second stage of \name{} then leverages this information to identify and describe the associated visual feature.
\item We introduce \bench{}, the first benchmark designed to evaluate automated methods for systematic misalignment detection. \bench{} consists of 420 image-caption datasets, each paired with a ground-truth label for a systematic misalignment. Methods are then quantitatively evaluated on the extent to which their predictions align with the ground truth. 
\vspace{-1mm}
\end{itemize}

We evaluate \name{} using \bench{}, analyzing a range of approaches for each subtask. The best configuration of \name{} correctly identifies the systematic misalignment in 63.8\% of \bench{} datasets. \name{} exhibits a nearly 4x improvement over the closest baseline, demonstrating the utility of our dual-stage, structured approach for addressing the systematic misalignment detection task. Finally, we supplement our evaluations on \bench{} with real-world evaluations, demonstrating quantitatively and qualitatively that (1) \name{} can accurately surface systematic misalignments in captions generated by four MLLMs and (2) \name{} is a powerful tool for auditing off-the-shelf datasets with MLLM-generated captions.

Ultimately, we envision our novel task, benchmark, and method aiding in the following real-world contexts. First, our approach reveals insights into failure modes of trained MLLMs, which can (1) provide developers with critical information for building more robust models as well as (2) assist end-users with understanding limitations prior to real-world deployment. For instance, returning to our previous example, physicians using an MLLM in the clinic can be forewarned that generated reports tend to incorrectly diagnose ``cardiomegaly" when X-rays have visible ``pacemakers"; knowledge of this failure mode can allow for further manual review of model outputs on those cases. Second, our approach can help users identify systematic captioning errors in off-the-shelf datasets, even \textit{in black-box settings where access to the underlying MLLM is unavailable}. This is a particularly important use-case, especially as publicly-available image datasets with MLLM-generated captions become widely used for training the next generation of multimodal foundation models.

\paragraph{Conflict of Interest Disclosure.} None. Funding sources are listed in the Acknowledgments at the end of this paper.
\section{Related Work}
\label{app:related_work}

Our work builds on three prior lines of study: (1) \textit{local misalignment detection} methods that identify captioning errors at the per-sample level; (2) \textit{global error detection} methods that summarize systematic trends in prediction errors; and (3) methods for \textit{describing patterns in large datasets with natural language}. 

\textbf{Local Misalignment Detection:} Given a single image and its paired model-generated caption, one line of recent work has focused on developing metrics that measure image-caption alignment using numeric scores. Examples include reference-free metrics like CLIPScore \citep{hessel-etal-2021-clipscore} and PAC-S \citep{Sarto_2023_CVPR_pacs}, which do not require the existence of ground-truth captions; on the other hand, reference-based metrics such as BLEU \citep{papineni-etal-2002-bleu}, ROUGE \citep{lin-2004-rouge}, CIDEr \citep{vedantam2015cider}, METEOR \citep{banerjee-lavie-2005-meteor}, and RefCLIPScore \citep{hessel-etal-2021-clipscore} make use of ground-truth captions. The utility of such metrics is typically evaluated using image-caption benchmarks with human-annotated quality judgments (e.g. FLICKR8K-Expert \citep{Hodosh2013flickr}, Pascal-50S \citep{vedantam2015cider}, ReXVal \citep{rexval}) or known model-injected errors (e.g. FOIL \citep{shekhar-etal-2017-foil}, ReXErr \citep{rexerr}). 

Several recent works have extended numeric scoring strategies by proposing interpretable metrics, which are capable of identifying the specific features in model-generated captions that are incorrect with respect to the image. Examples include reference-based metrics like CHAIR \citep{rohrbach-etal-2018-chair}, ALOHa \citep{petryk-etal-2024-aloha}, and GREEN \citep{ostmeier-etal-2024-green} as well as reference-free metrics like FLEUR \citep{lee-etal-2024-fleur}. Our work draws inspiration from these studies by also prioritizing interpretability; our method \name{} not only detects whether captioning errors are present but also provides users with a natural language output indicating the erroneous textual facts and associated visual cues. However, our study exhibits a key distinction from this line of work: whereas these metrics evaluate a single image and its paired model-generated caption, our work instead focuses on detecting \textit{global}, systematic trends in captioning errors.

\textbf{Global Error Detection:} Due to visual biases or spurious correlations learned during training, machine learning models often make systematic prediction errors at test time. Selected examples in the classification setting noted by prior works include (1) an object recognition model that can correctly classify cows in pastoral settings yet demonstrates high error rates when cows are in beach settings \citep{Beery_2018_ECCV} and (2) a pneumothorax detection model that achieves radiologist-level overall accuracy yet demonstrates high error rates when chest tubes, a medical device used for treatment, are absent \citep{hiddenstratification}. Detecting such failures is challenging due to the fact that relevant subgroups are typically not annotated in data.

A recent line of work has explored the development of automated methods for identifying global, systematic error patterns in classification settings. Given a validation dataset with images, model predictions, and ground-truth labels, these methods identify specific visual features (e.g. the beach background or the absence of tubes in the above examples) that are associated with higher error rates \citep{domino2022,jain2023distilling,sohoni2020george,varma2024ravldiscoveringmitigatingspurious}. Our work shares a similar goal in identifying systematic error patterns; however, we extend beyond the classification setting to the image captioning setting, where input datasets consist of images and paired model-generated captions. The inclusion of free-form text in input datasets presents an added level of complexity in comparison to labels. Additionally, we explicitly consider settings where ground-truth captions are unavailable.

\textbf{Describing Datasets with Natural Language:} Several works have presented approaches for describing patterns in large datasets using natural language \citep{burgess2025videoactiondifferencing}. In particular, recent studies have generated natural language descriptions (i) summarizing differences given two input datasets \citep{VisDiff,zhong2022describingdifferencestextdistributions} and (ii) summarizing model prediction errors given classification datasets with labels \citep{domino2022,r-menon-srivastava-2024-discern,Kim_2024_CVPR}. Our work also involves summarizing dataset-level patterns with natural language; however, in our setting, datasets consist of images and paired captions, and descriptions must specifically identify systematic misalignments.
\section{Task Definition}
\label{sec:taskdefinition}

In this section, we formally introduce the systematic misalignment detection task. Consider a vision-language dataset $\mathcal{D} = \{(V_i, T_i)\}_{i=1}^N$ consisting of images $V$ paired with free-form, model-generated text $T$. For example, dataset $\mathcal{D}$ may consist of chest X-rays $V$ paired with MLLM-generated radiology reports $T$. We will express each text sample $T_i$ as a collection of textual facts $T_i = \{t^i_1,t^i_2,...,t^i_{n_i}\}$ and each image $V_i$ as a collection of visual features $V_i = \{v^i_1,v^i_2,...,v^i_{m_i}\}$. 

Dataset $\mathcal{D}$ may include misaligned samples, where text $T_i$ does not accurately describe the content of the paired image $V_i$. We consider a pair $(V_i,T_i)$ to be misaligned if there exists at least one erroneous textual fact $t^i_k \in T_i$ that does not accurately describe any visual feature $v^i_j \in V_i$. Misalignments are particularly egregious when they occur in a \textit{systematic} fashion, meaning that an erroneous textual fact $t$ is repeatedly associated with the presence of a visual feature $v$ throughout a dataset. For instance, in the medical imaging example discussed earlier, incorrect diagnoses of cardiomegaly in MLLM-generated reports are strongly associated with the presence of a pacemaker in the corresponding chest X-rays; this suggests the existence of a systematic misalignment between reports containing 
$t=cardiomegaly$ and images containing $v=pacemaker$.

Thus, given $\mathcal{D}$, the goal of the \textbf{systematic misalignment detection} task is to discover textual errors $t$ that are systematically associated with visual cues $v$. A method $\mathcal{M}: \mathcal{D} \rightarrow (\hat{t}, \hat{v})$ that aims to solve the systematic misalignment detection task will accept dataset $\mathcal{D}$ as input; we note here that datasets may be large in size, consisting of thousands of image-text pairs. Then, method $\mathcal{M}$ will predict ($\hat{t}$, $\hat{v}$) as output, indicating the discovered textual error $\hat{t}$ and associated visual feature $\hat{v}$; here, both $\hat{t}$ and $\hat{v}$ will be expressed in text. 

We consider two variants of input dataset $\mathcal{D}$: (1) \textit{reference-free}, where each sample in dataset $\mathcal{D} = \{(V_i, T_i)\}_{i=1}^N$ consists of image $V_i$ and model-generated text $T_i$, and (2) \textit{reference-based}, where each sample in dataset $\mathcal{D} = \{(V_i, T_i, R_i)\}_{i=1}^N$ consists of an image $V_i$, model-generated text $T_i$, and a ground-truth reference caption $R_i$. 

\section{Our Approach: \name{}}

\begin{figure}[t]
\begin{center}
\includegraphics[width=0.8\linewidth]{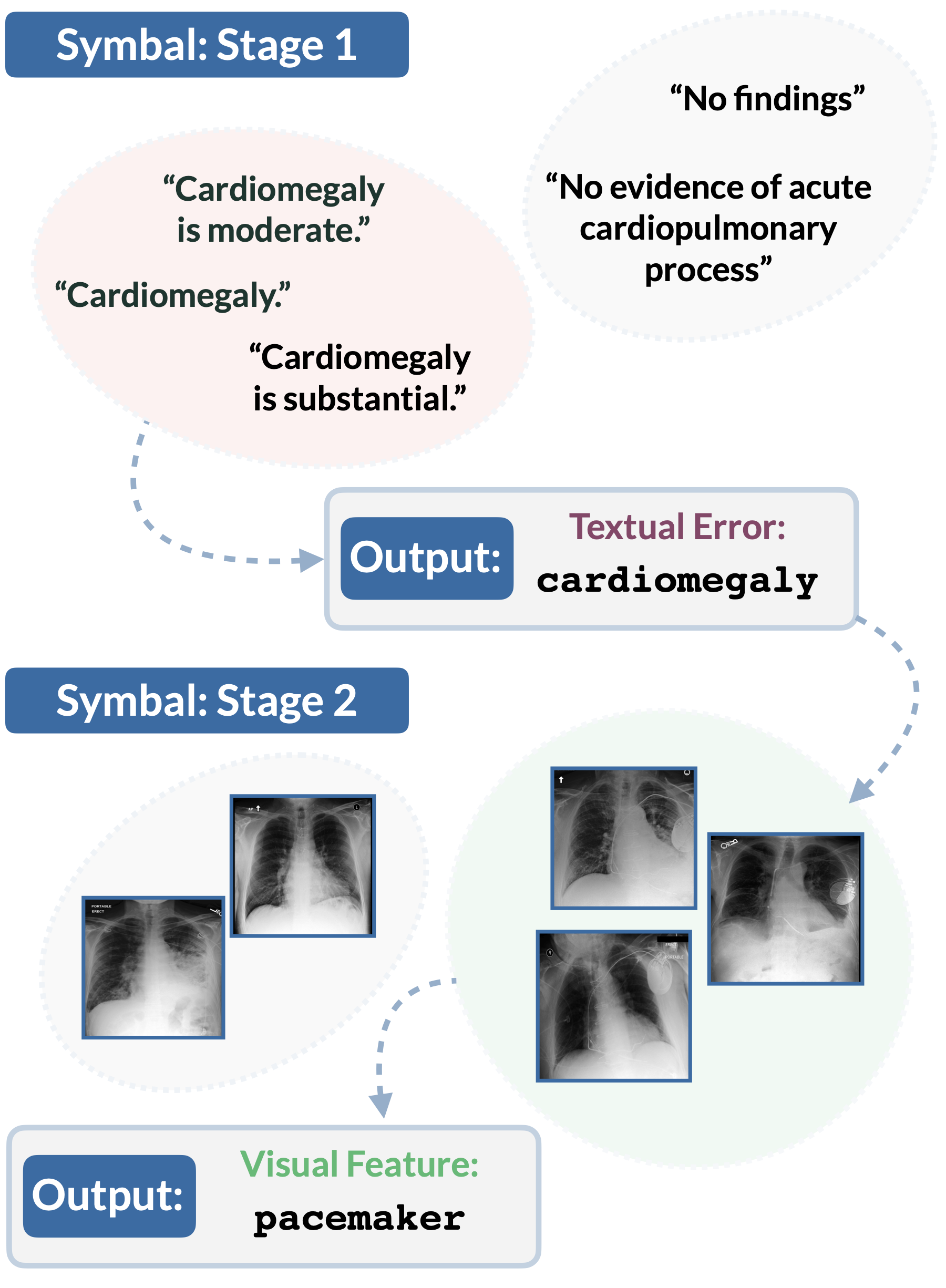}
\end{center}
  \caption{\name{} detects systematic misalignments with a two-stage procedure. The first stage involves detecting erroneous textual facts, and the second stage involves detecting associated visual features.}
\label{fig:methodfig}
\end{figure}

The systematic misalignment detection task is made challenging by the fact that vision-language datasets may be complex and large in size; identifying global error patterns from such datasets is nontrivial. In this section, we address this challenge with our approach \name{}, which structures the systematic misalignment detection task into two stages. Each stage is comprised of three individual subtasks: grouping, scoring, and summarizing. Sections \ref{sec:methodstage1} and \ref{sec:methodstage2} discuss the two stages in detail.

\subsection{Stage 1: Detecting Erroneous Textual Facts}
\label{sec:methodstage1}

The first stage of \name{} predicts the erroneous textual fact by (1) grouping semantically-similar facts that occur consistently throughout the dataset, (2) scoring each group of facts by degree of misalignment with paired images, (3) and summarizing the top-ranked group of facts into a single unifying concept $\hat{t}$. The three subtasks associated with Stage 1 are detailed below:

\begin{itemize}[leftmargin=*]
    \item \textbf{Grouping semantically-similar facts:} As defined in Section \ref{sec:taskdefinition}, we first express each text sample $T_i$ as a collection of textual facts $T_i = \{t^i_1,t^i_2,...,t^i_{n_i}\}$ by splitting captions at the sentence level. We then identify clusters of semantically-similar facts that occur in $\mathcal{D}$; for example, in the medical imaging example discussed earlier, perhaps one such cluster will contain sentences from radiology reports that discuss the presence of cardiomegaly. To this end, we aggregate all textual facts in $\mathcal{D}$, forming the set $\bigcup_{i=1}^N T_i = \{t_k^i: i=1,...,N; k=1,...,n_i\}$. Each textual fact in this set is encoded using a text embedding model; then, embeddings are clustered using spherical K-Means, where the number of clusters is selected automatically using Silhouette distance.
    \item \textbf{Scoring groups by degree of misalignment:} Next, we score each cluster by computing the mean degree of alignment between constituent textual facts and paired images. Based on methods from prior work \citep{hessel-etal-2021-clipscore,VisDiff,chen2024mllmasajudgeassessingmultimodalllmasajudge}, we consider three options for measuring alignment between a given textual fact and its paired image: (1) \textit{embedding scorer}, which computes embeddings for the text and image modalities and measures alignment as the cosine similarity, (2) \textit{text-only scorer}, which generates a caption for the image and tasks an LLM with determining if the textual fact is accurate with respect to the caption, and (3) \textit{vision-language scorer}, where a MLLM is provided both the image and the textual fact as input and tasked with determining if the textual fact is accurate. Low scores suggest that a large proportion of textual facts in the cluster are misaligned with respect to their paired images.
    \item \textbf{Summarizing the top-ranked group:} Given the alignment scores computed in the previous step, we identify the cluster exhibiting the highest degree of misalignment, which we refer to as $C_{text}$. Then, we apply a \textit{text-only summarizer}, where an LLM is provided a list of textual facts in $C_{text}$ and tasked with identifying the unifying concept.
\end{itemize}

The final output of the summarizer is the predicted erroneous textual fact $\hat{t}$; for example, in the medical example discussed earlier, the predicted textual fact may be $\hat{t} = cardiomegaly$. In Section \ref{subsec:textresults}, we evaluate the role of various text embedding models and alignment scorers.

\subsection{Stage 2: Detecting Associated Visual Features}
\label{sec:methodstage2}

We now proceed to the second stage of \name{}, which predicts the associated visual feature by (1) grouping semantically-similar images paired with text containing fact $\hat{t}$, (2) scoring each group of images by degree of misalignment with $\hat{t}$, and (3) summarizing the top-ranked group of images into a single unifying concept $\hat{v}$. The three subtasks associated with Stage 2 are detailed below:

\begin{itemize}[leftmargin=*]
    \item \textbf{Grouping semantically-similar images:} We begin by identifying all images $V_i \in \mathcal{D}$ containing at least one paired textual fact in cluster $C_{text}$ (i.e. where $t^i_k \in C_{text}$ for some $k$). Each image in this set is encoded using an image embedding model; then, embeddings are clustered using spherical K-Means, where the number of clusters is selected automatically using Silhouette distance. 
    \item \textbf{Scoring groups by degree of misalignment:} Next, we score each cluster by computing the mean degree of misalignment between images and paired textual facts in $C_{text}$. We consider the same scoring mechanisms as in Stage 1. Low scores suggest that a large proportion of images in the cluster are misaligned with fact $\hat{t}$.
    \item \textbf{Summarizing the top-ranked group:} Given the alignment scores computed in the previous step, we identify the cluster exhibiting the highest degree of misalignment, which we will refer to as $C_{image}$. Then, we consider two summarization mechanisms for identifying the unifying concept shared by images in $C_{image}$:  (1) \textit{text-only summarizer}, where a caption is generated for each image in $C_{image}$ and an LLM is tasked with identifying the unifying concept, and (2) \textit{vision-language summarizer}, where an MLLM is provided with images in $C_{image}$ and tasked with identifying the unifying concept.  
\end{itemize}

The final output of the summarizer is the predicted visual feature $\hat{v}$; for example, in the medical example discussed earlier, the predicted visual feature may be $\hat{v} = pacemaker$. In Section \ref{subsec:imageresults}, we evaluate the role of various image embedding models, alignment scorers, and summarizers.

We note here that some datasets may contain multiple systematic misalignments; \name{} can be trivially extended to such settings, as we show in Appendix \ref{appendixsec:symbaldetails} and \ref{appendixsec:realworld}.
\section{Benchmark: \bench{}}
\label{sec:benchmark}
The key challenge behind evaluating methods like \name{} on real-world vision-language datasets is that ground-truth systematic misalignments are typically unknown. Moreover, collecting human annotations for a task at this scale, where datasets include thousands of images paired with information-dense captions, is simply intractable. Thus, without access to ground-truth annotations, it becomes difficult (1) to determine whether misalignments identified by a method like \name{} are accurate and (2) to quantitatively compare results across multiple methods.

 In this section, we introduce \bench{}, which is designed to address this challenge. Specifically, \bench{} utilizes an automated method to inject a pre-defined systematic misalignment into a base vision-language dataset, yielding an evaluation setting where a ground-truth annotation ($t$, $v$) is available. The automated nature of our approach provides several key advantages, including (1) the ability to generate hundreds of evaluation settings simply by injecting varied systematic misalignments, (2) the presence of ground-truth labels that are guaranteed to be accurate, and (3) the ability to extend to specialized domains like medical imaging. In Section \ref{subsec:realworld}, we augment our evaluations on \bench{} with real-world analyses.

\textbf{Benchmark Design:} \bench{} consists of 420 \textit{evaluation settings}, where each setting is comprised of a vision-language dataset $\mathcal{D}$ and an associated ground-truth label ($t$,$v$) representing the systematic misalignment. In order to create each evaluation setting, we (1) obtain a high-quality base dataset with images and paired text, (2) predefine a systematic misalignment ($t$, $v$), and (3) inject the erroneous textual fact $t$ into the base dataset such that a strong association exists with visual feature $v$. Below, we discuss these three steps in detail: 
\begin{enumerate}[leftmargin=*]
\item \textbf{Obtaining a base dataset.} We begin by obtaining an off-the-shelf vision-language dataset with high-quality samples. We consider two options for the base dataset: COCO (2017 val split) \citep{lin2015microsoftcococommonobjects} and MIMIC-CXR (test split) \citep{johnson2019mimiccxrjpglargepubliclyavailable}. COCO consists of natural images depicting common objects from 80 categories. After preprocessing, the base dataset includes a total of 4349 images with associated captions. MIMIC-CXR consists of chest X-rays and associated radiology reports obtained from the Beth Israel Deaconess Medical Center. After preprocessing, the base dataset includes 2233 images, each paired with the  ``Impressions" section of the corresponding report. 

\item \textbf{Predefining a systematic misalignment.} Given a base dataset, we predefine a systematic misalignment consisting of a textual fact $t$ and associated visual feature $v$. Predefined misalignments are meant to emulate those that are likely to emerge when using real-world, off-the-shelf MLLMs to generate captions. For COCO, we sample $t$ and $v$ from the set of 80 object categories present in the dataset. For MIMIC-CXR, we sample $t$ from a set of five disease categories (cardiomegaly, pneumothorax, atelectasis, pleural effusion, and edema) and $v$ from a set of five medical devices (pacemaker, chest tube, endotracheal tube, surgical clips, sternotomy wires).\footnote{We define these options for $t$ and $v$ due to the fact that medical imaging models often learn spurious associations between medical devices and disease categories, as documented in prior work (e.g. \cite{hiddenstratification}); thus, our predefined misalignments are highly plausible in real-world, model-generated reports.} 

\item \textbf{Injecting the predefined systematic misalignment.} We insert the erroneous textual fact $t$ into text samples in the base vision-language dataset such that a strong association exists between text containing $t$ and images containing visual feature $v$. The strength of the association is controlled using Cramer's V scores. Each inserted fact $t$ is formatted as a sentence using diverse templates. 
\end{enumerate}

We repeat this procedure across the two possible options for the base dataset and a range of possible options for $t$ and $v$, yielding 420 evaluation settings encompassing a total of 1.7 million image-text pairs. Additional details are in Appendix \ref{appendixsec:benchdetails} and \ref{appendixsec:descriptivestats}.

\textbf{Benchmark Evaluation:} We will use the notation $\{(\mathcal{D}_s,(t_s,v_s))\}_{s=1}^{420}$ to represent \bench{}, where the evaluation setting with index $s$ has an associated dataset $\mathcal{D}_s$ and ground-truth label $(t_s,v_s)$. We construct both reference-based and reference-free variants of \bench{}, which differ only with respect to whether $\mathcal{D}_s$ includes reference captions. At evaluation time, dataset $\mathcal{D}_s$ will be provided to method $\mathcal{M}$, which will output a prediction $(\hat{t}_s,\hat{v}_s)$. We count the prediction as accurate if the top-K predictions for $\hat{t}_s$ include $t_s$ and the top-K predictions for $\hat{v}_s$ include $v_s$. Here, we evaluate equivalence using LLM-as-a-Judge with Llama3.3-70B \citep{grattafiori2024llama3herdmodels}. Overall performance on \bench{} is measured with Accuracy@K, computed as the percentage of the 420 settings in \bench{} where the prediction is accurate.
\section{Results}
\label{sec:results}

We now evaluate \name{} on the systematic misalignment detection task. In Sections \ref{subsec:textresults} and \ref{subsec:imageresults}, we use \bench{} to analyze the choice of embedding models, alignment scorers, and summarizers. In Section \ref{subsec:endtoendresults}, we perform end-to-end evaluations of the best configuration of \name{}, comparing with baselines and performing fine-grained analyses. Finally, in Section \ref{subsec:realworld}, we extend beyond \bench{} to real-world settings.

\begin{table*}[t]
\caption{We evaluate various text embedding models, alignment scorers, and summarizers on the performance of \name{} Stage 1.}
\centering
\resizebox{\textwidth}{!}{ 
\begin{tabular}{clll|cc|cc}
\toprule
&
& 
& 
& \multicolumn{2}{c|}{ \textbf{Reference-Free}}
& \multicolumn{2}{c}{ \textbf{Reference-Based}}
\\
& \textbf{\small Text Embedding} & \textbf{\small Alignment Scorer} & \textbf{\small Summarizer}  & Acc@1 & Acc@5 & Acc@1 & Acc@5\\
\midrule
\parbox[t]{2mm}{\multirow{4}{*}{\rotatebox[origin=c]{90}{Natural}}} & Qwen3-8B & Vision-Language (Qwen-72B) & Text-Only (Qwen-72B)  & \textbf{92.8} & \textbf{94.2} & 80.8 & 82.8\\
& OpenCLIP & Vision-Language (Qwen-72B) & Text-Only (Qwen-72B) & \textbf{92.8} & 93.9 & \textbf{86.1} & \textbf{87.8}\\
& Qwen3-8B & Text-Only (Qwen-72B) & Text-Only (Qwen-72B)  & 82.8 & 85.0  & 81.9 & 83.9\\
& OpenCLIP & Text-Only (Qwen-72B) & Text-Only (Qwen-72B) & 64.2 & 67.2  & 67.5 & 71.4\\
\midrule
\parbox[t]{2mm}{\multirow{4}{*}{\rotatebox[origin=c]{90}{Medical}}} & XRayCLIP & Text-Only (MedGemma-27B) & Text-Only (MedGemma-27B) & \textbf{51.7} & \textbf{75.0}  & 88.3 & 95.0\\
& XRayCLIP & Text-Only (MedGemma-27B) & Text-Only (Qwen-72B) & \textbf{51.7} & 73.3  & \textbf{100.0} & \textbf{100.0}\\
& XRayCLIP & Text-Only (Qwen-72B) & Text-Only (MedGemma-27B) & 26.7 & 58.3  & 90.0 & 93.3\\
& MedSigLIP & Text-Only (MedGemma-27B) & Text-Only (MedGemma-27B) & 30.0 & 53.3  & 83.3 & \textbf{100.0}\\
\bottomrule
\end{tabular}
}
\label{table:textonly}
\end{table*}

\begin{table*}[t]
\caption{We evaluate various image embedding models, alignment scorers, and summarizers on the performance of \name{} Stage 2.}
\centering
\resizebox{\textwidth}{!}{ 
\begin{tabular}{clll|cc|cc}
\toprule
&
& 
& 
& \multicolumn{2}{c|}{ \textbf{Reference-Free}}
& \multicolumn{2}{c}{ \textbf{Reference-Based}}
\\
& \textbf{\small Image Embedding} & \textbf{\small Alignment Scorer} & \textbf{\small Summarizer} & Acc@1 & Acc@5 & Acc@1 & Acc@5\\
\midrule
\parbox[t]{2mm}{\multirow{4}{*}{\rotatebox[origin=c]{90}{Natural}}}  & OpenCLIP & Vision-Language (Qwen-72B) & Text-Only (Qwen-72B) & \textbf{49.7 }& \textbf{69.7} & 41.9 & 52.2\\
& OpenCLIP & Embedding (OpenCLIP) & Vision-Language (Qwen-72B) & 48.1 & 63.9 & 42.5 & 55.6\\
& OpenCLIP & Embedding (OpenCLIP) & Text-Only (Qwen-72B) & 47.8 & 62.8 & \textbf{43.9} & \textbf{55.8}\\
& OpenCLIP & Vision-Language (Qwen-72B) & Vision-Language (Qwen-72B) & 45.8 & 62.5 & 38.9 & 52.2\\
\midrule
\parbox[t]{2mm}{\multirow{4}{*}{\rotatebox[origin=c]{90}{Medical}}} & XRayCLIP & Embedding (MedSigLIP) & Vision-Language (MedGemma-27B)  & 11.7 & \textbf{36.7}  & 28.3 & 53.3\\
& MedSigLIP & Embedding (MedSigLIP) & Vision-Language (MedGemma-27B)  & 11.7 & 31.7  & 25.0 & 46.7\\
& OpenCLIP & Embedding (MedSigLIP) & Vision-Language (MedGemma-27B)  & \textbf{13.3} & 28.3 & 20.0 & 46.7\\
& MedSigLIP & Embedding (XRayCLIP) & Vision-Language (MedGemma-27B) & 10.0 & 28.3  & \textbf{33.3} & \textbf{60.0}\\
\bottomrule
\end{tabular}
}
\label{table:imageonly}
\vspace{-0.1in}
\end{table*}

\subsection{\name{} Detects Erroneous Textual Facts}
\label{subsec:textresults}

We first evaluate the role of various text embedding models, alignment scorers, and summarizers on the performance of Stage 1 of \name{}, which aims to predict the erroneous textual fact $\hat{t}_s$ given an input dataset $\mathcal{D}_s$ in \bench{}. We compute Accuracy@1 and Accuracy@5 by comparing $\hat{t}_s$ with $t_s$ across all 420 settings in \bench{}. 
Results are summarized in Table \ref{table:textonly}.

For the natural image datasets in \bench{}, Table \ref{table:textonly} Upper demonstrates the performance of the top-four compositions, ranked by Accuracy@5 scores on the reference-free setting. Our results show that the best-performing variant of \name{} (shown in Row 1 of Table \ref{table:textonly} Upper) achieves strong performance, correctly identifying the erroneous textual fact in 94.2\% (Acc@5) of \bench{} datasets in the reference-free configuration and 82.8\% (Acc@5) of \bench{} datasets in the reference-based configuration. Interestingly, we find that performance in reference-free settings is often substantially higher than performance in the reference-based setting, which is likely a result of the sparse information content often present in COCO reference captions. When considering the composition of \name{}, we note that the choice of the alignment scorer appears to be most important; the vision-language scorer substantially outperforms the text-only scorer with the same underlying model (Qwen2.5-72B). 

Given these results, we select the Qwen3-Embedding-8B text embedding model \citep{qwen3embedding}, the vision-language alignment scorer with Qwen2.5-72B \citep{qwen2025qwen25technicalreport}, and the text-only summarizer with Qwen2.5-72B \citep{qwen2025qwen25technicalreport} for all future \name{} evaluations on natural images.

For the medical image datasets in \bench{}, Table \ref{table:textonly} Lower demonstrates the performance of the top-four compositions. Our results show that the best-performing variant of \name{} (shown in Row 1 of Table \ref{table:textonly} Lower) correctly identifies the erroneous textual feature in 75.0\% (Acc@5) of datasets in the reference-free configuration and 95.0\% (Acc@5) of datasets in the reference-based configuration. In contrast to the natural image datasets, we find that the reference-free configuration is harder than the reference-based configuration, likely due to the complexity of medical image data; alignment scoring in this domain is challenging without access to reference text. We also note that a key advantage of \name{} is its ability to extend to specialized domains simply by interchanging constituent models with domain-specific versions. 

Given these results, we select the XRayCLIP-ViT-L text embedding model \citep{chexagent-2024}, the text-only alignment scorer with MedGemma-27B \citep{sellergren2025medgemmatechnicalreport}, and the text-only summarizer with MedGemma-27B \citep{sellergren2025medgemmatechnicalreport} for all future \name{} evaluations on medical images.

\subsection{\name{} Detects Associated Visual Features}
\label{subsec:imageresults}

We next evaluate the role of various image embedding models, alignment scorers, and summarizers on the performance of Stage 2 of \name{}. We hold the composition of Stage 1 constant using results from Section \ref{subsec:textresults}. We compute Accuracy@1 and Accuracy@5 by comparing $\hat{v}_s$ with $v_s$ across all 420 settings in \bench{}. Results are summarized in Table \ref{table:imageonly}.

For the natural image datasets in \bench{}, Table \ref{table:imageonly} Upper demonstrates the performance of the top-four compositions, ranked by Accuracy@5 scores on the reference-free setting. Our results show that the best-performing variant of \name{} (shown in Row 1 of Table \ref{table:imageonly} Upper) correctly identifies the visual feature in 69.7\% (Acc@5) of datasets in the reference-free configuration and 52.2\% (Acc@5) of datasets in the reference-based configuration. We observe that performance values in Table \ref{table:imageonly} are lower than Table \ref{table:textonly}, suggesting that identifying visual features that systematically occur with textual errors is substantially more challenging than identifying the textual error itself. We also observe that the best-performing variant of \name{} utilizes the same alignment scorer and summarizer as in Stage 1. 

Given these results, we select the OpenCLIP-ViT-H image embedding model \citep{ilharco_gabriel_2021_5143773}, vision-language alignment scorer with Qwen2.5-72B \citep{qwen2025qwen25technicalreport}, and text-only summarizer with Qwen2.5-72B \citep{qwen2025qwen25technicalreport} for all future \name{} evaluations on natural images.

For the medical image datasets in \bench{}, Table \ref{table:imageonly} Lower demonstrates the performance of the top-four compositions, ranked by Accuracy@5 scores on the reference-free setting. Our results show that the best-performing variant of \name{} (shown in Row 1 of Table \ref{table:imageonly} Lower) correctly identifies the visual feature in 36.7\% (Acc@5) of datasets in the reference-free configuration and 53.3\% (Acc@5) of datasets in the reference-based configuration. Our results suggest that identifying visual features in the medical domain is a particularly challenging task in both reference-free and reference-based settings, and consequently, the optimal composition of alignment scorers and summarizers differs markedly from those identified in Stage 1. 

Given these results, we select the XRayCLIP-ViT-L image embedding model \citep{chexagent-2024}, embedding alignment scorer with MedSigLIP \citep{sellergren2025medgemmatechnicalreport}, and vision-language summarizer with MedGemma-27B \citep{sellergren2025medgemmatechnicalreport} for future evaluations on medical images.

\subsection{\name{} Shows Strong End-to-End Performance}
\label{subsec:endtoendresults}

\begin{figure}[t]
\begin{center}
\includegraphics[width=0.95\linewidth]{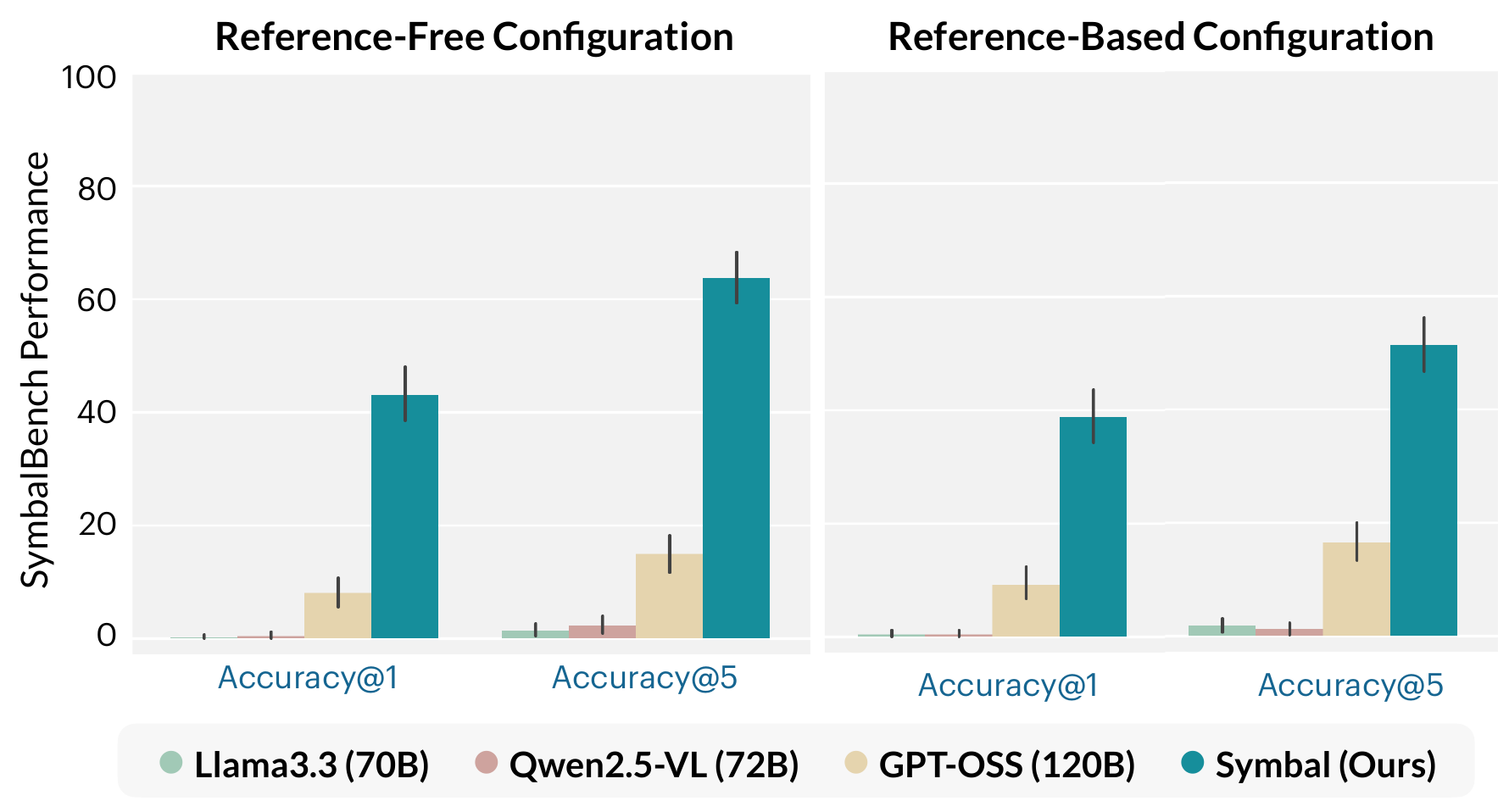}
\end{center}
\vspace{-0.05in}
  \caption{\name{} demonstrates strong end-to-end performance on \bench{}, substantially outperforming baselines.}
\label{fig:end2end}
\vspace{-0.1in}
\end{figure}

\begin{figure}[t]
\begin{center}
\includegraphics[width=\linewidth]{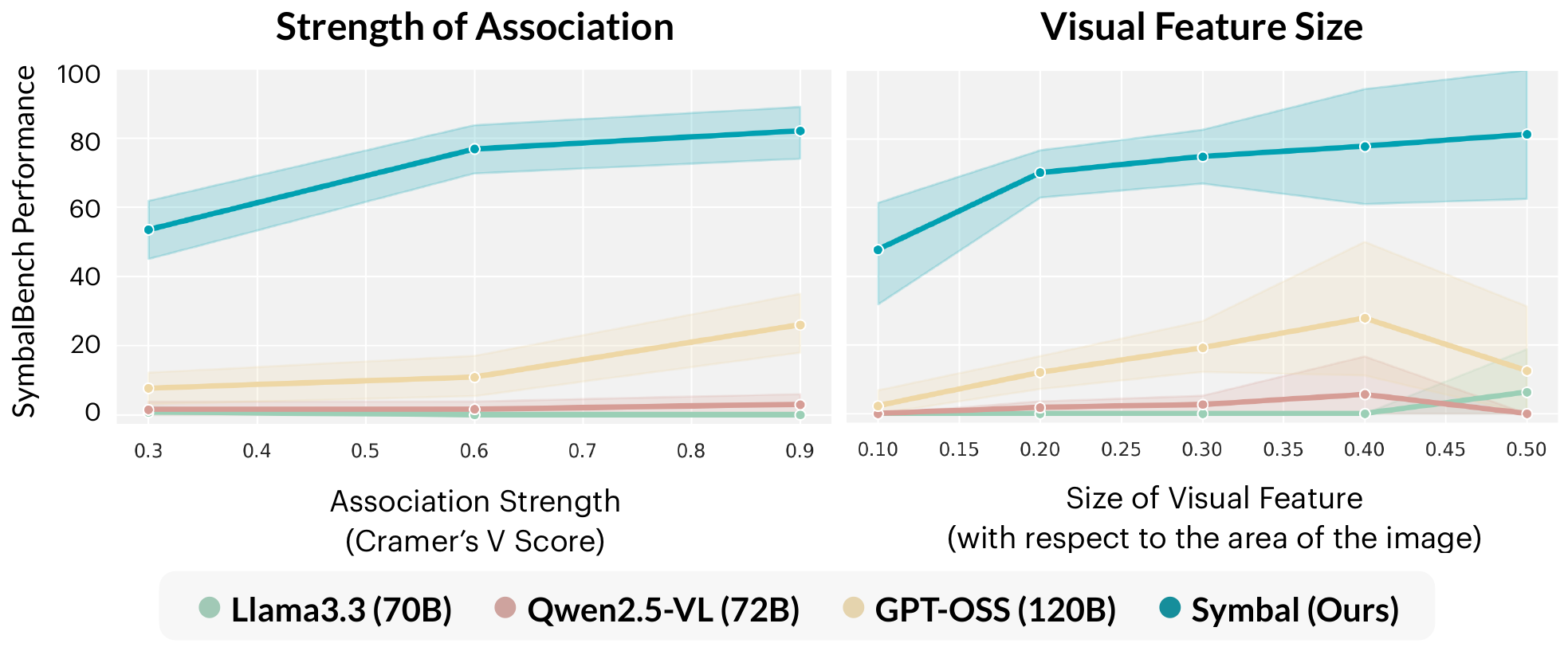}
\end{center}
\vspace{-0.05in}
  \caption{We report performance on \bench{} (reference-free) stratified across association strengths and visual feature sizes. This analysis focuses on natural image settings in \bench{}.}
\label{fig:finegrained}
\vspace{-0.15in}
\end{figure}

Given an optimal composition of \name{}, we now perform end-to-end analyses across \bench{}. Since our study proposes a novel task, there are no existing baselines for comparison. As a result, we compare the structured, dual-stage approach of \name{} to a single-stage, direct-prompting method where each dataset $\mathcal{D}_s$ is directly provided to an off-the-shelf LLM in the form of a text prompt; the LLM is then instructed to output the erroneous textual fact and the associated visual feature. Three state-of-the-art LLMs are considered (i.e. Llama3.3 70B, Qwen2.5-VL 72B, and GPT-OSS 120B), selected to ensure a fair comparison with \name{} due to comparable parameter counts. As the token length of the direct prompts far surpasses the context window of these LLMs, we use only a sample of each dataset, ensuring that the final inference procedure requires no more compute resources than \name{}.

In Figure \ref{fig:end2end}, we measure the extent to which \name{} can accurately predict \textit{both} the textual fact $\hat{t}_s$ and the visual feature $\hat{v}_s$ across both the reference-free and reference-based variants of \bench{}. Results show that the systematic misalignment detection task is highly challenging in both experimental settings, with several baselines generating few correct predictions. \name{} successfully identifies the systematic misalignment in up to 63.8\% of datasets in \bench{}, with the highest performance observed in the reference-free setting (Accuracy@5). \name{} outperforms the closest baseline (GPT-OSS 120B) across all experimental settings, with GPT-OSS 120B correctly identifying the misalignment in only 17.1\% of \bench{} datasets in the best case. These results demonstrate that the structured, dual-stage approach utilized by \name{} provides substantial performance benefits over single-stage, direct prompting baselines.

In Figure \ref{fig:finegrained}, we provide a stratified breakdown of \name{} performance. \name{} outperforms baselines across highly-challenging subsets of \bench{} where (1) the strength of the systematic misalignment is weak (i.e. weak association between the textual error and visual feature as measured by Cramer's V scores) and (2) visual features are small in size.

Extended results and ablations are provided in Appendix Section \ref{appendixsec:results}.

\subsection{\name{} Extends to Real-World Settings}
\label{subsec:realworld}

\begin{figure*}[t]
\begin{center}
\includegraphics[width=0.95\linewidth]{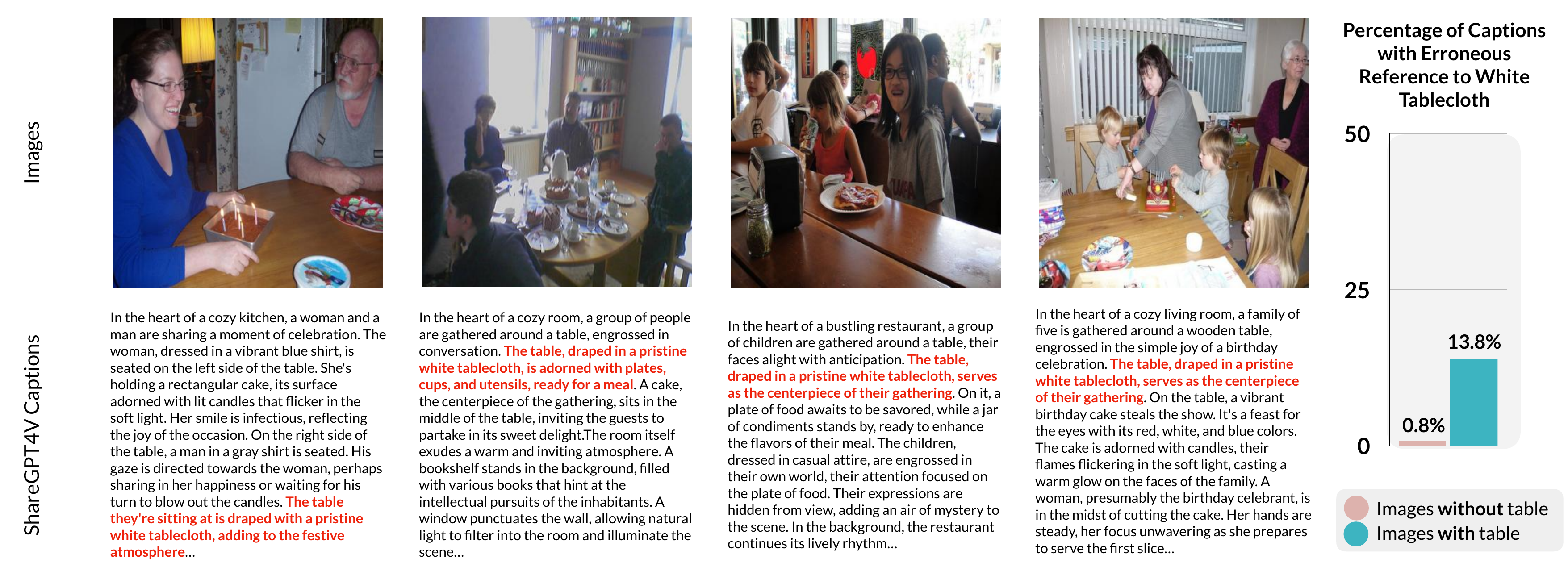}
\end{center}
\vspace{-0.05in}
  \caption{\name{} discovers systematic misalignments in ShareGPT4V, an off-the-shelf dataset with model-generated captions.}
\label{fig:sharegptxample_main}
\vspace{-0.2in}
\end{figure*}

In this section, we further demonstrate the utility of \name{} by supplementing our evaluations on \bench{} with additional quantitative and qualitative analyses in real-world settings. Our results show that (1) \name{} can accurately surface systematic misalignments in captions generated by off-the-shelf MLLMs and (2) \name{} is a powerful tool for auditing vision-language datasets.

\textbf{\name{} can accurately surface systematic misalignments in captions generated by off-the-shelf MLLMs.} First, we use \name{} to analyze captions generated by four real-world off-the-shelf MLLMs: Llava1.5-7B \citep{liu2023improvedllava}, Llava1.5-13B \citep{liu2023improvedllava}, AyaVision-8B \citep{dash2025ayavisionadvancingfrontier}, and LlavaOneVision-7B \citep{li2024llavaonevisioneasyvisualtask}. We utilize each model to generate captions for the COCO dataset (2017 val split); we then apply \name{} (reference-free) to predict systematic misalignments ($\hat{t}$, $\hat{v}$). 

As discussed in Section \ref{sec:benchmark}, evaluating predictions in real-world settings is highly challenging since ground-truth systematic misalignments are unknown. Here, in order to address this issue, we validate identified systematic misalignments in two ways. First, we \textit{qualitatively} validate the existence of \name{}-identified systematic misalignments with visual analysis. Second, we \textit{quantitatively} validate whether a link between erroneous fact $\hat{t}$ and visual feature $\hat{v}$ truly exists; to this end, we measure whether model-generated captions are indeed more likely to include erroneous references to $\hat{t}$ when $\hat{v}$ is present compared to when $\hat{v}$ is absent. In order to perform this evaluation, we use a state-of-the-art open-set object detector \citep{minderer2024scalingopenvocabularyobjectdetection} to annotate the presence of $\hat{v}$ in each image, and we use our top-performing alignment scorer (vision-language scorer with Qwen-72B) to annotate erroneous references to $\hat{t}$ in each caption. In Appendix \ref{appendixsec:realworld}, we demonstrate that automated annotations align closely with human judgments.

 \name{} identifies several systematic misalignments. In captions generated by Llava1.5-7B, \name{} detects that erroneous references to a \texttt{handbag} or a \texttt{handbag on the ground} ($\hat{t}$) in captions are often systematically associated with the presence of a \texttt{bus} ($\hat{v}$) in a scene, as shown in Figure \ref{appendixfig:realworld} [Row 2]. Quantitatively, our analysis finds that erroneous references to a \texttt{handbag} in model-generated captions are indeed 3.1 times more likely when a \texttt{bus} is present in the image compared to when a \texttt{bus} is absent, validating the \name{} prediction. In captions generated by LlavaOneVision-7B, \name{} detects that erroneous references to \texttt{text} ($\hat{t}$) in captions are often systematically associated with the presence of a \texttt{sign} ($\hat{v}$) in a scene, as shown in Figure \ref{appendixfig:realworld2} [Row 2]. This finding suggests that LlavaOneVision-7B struggles with OCR capabilities, where the presence of text-based signage in an image is likely to result in errors in the generated caption. Quantitatively, our analysis finds that erroneous references to \texttt{text} in model-generated captions are indeed 4.6 times more likely when a \texttt{sign} is present in the image compared to when a \texttt{sign} is absent, validating the \name{} prediction. Additional examples can be found in Appendix  \ref{appendixsec:realworld}.
 
\textbf{\name{} is a powerful tool for auditing open-source vision-language datasets.} Second, we use \name{} to analyze ShareGPT4V, an open-source image dataset with MLLM-generated captions commonly used as a pretraining dataset for vision-language models \citep{chen2023sharegpt4vimprovinglargemultimodal}. We sample a subset of 10k image-caption pairs from the ShareGPT4V dataset, and we then apply \name{} (reference-free) to predict systematic misalignments ($\hat{t}$, $\hat{v}$). Here, \name{} detects that erroneous references to a \texttt{white tablecloth} ($\hat{t}$) in captions are often systematically associated with the presence of a \texttt{table}, \texttt{cake}, and/or \texttt{people} ($\hat{v}$) in the scene, as shown in Figure \ref{fig:sharegptxample_main}. Quantitatively, our analysis finds that erroneous references to a \texttt{white tablecloth} in model-generated captions are indeed 17.2 times more likely when a \texttt{table} is present in the image compared to when a \texttt{table} is absent, validating the \name{} prediction. Additional examples are provided in Appendix \ref{appendixsec:realworld}.

As large-scale datasets like ShareGPT4V become increasingly prevalent, it becomes critical for users to be aware of potential systematic misalignments, as these errors can propagate to trained models. Specifically, if a dataset contains a systematic misalignment between erroneous textual fact $\hat{t}$ and visual feature $\hat{v}$, models trained on the dataset are likely to learn spurious correlations between $\hat{t}$ and $\hat{v}$, leading to prediction errors at test-time \citep{varma2024ravldiscoveringmitigatingspurious}. \name{} can aid users with understanding limitations of datasets with MLLM-generated captions as well as assist model developers with improving performance of MLLMs.
\section{Discussion}

In this work, we introduce the systematic misalignment detection task, which aims to identify textual errors in MLLM-generated captions that are systematically associated with visual features. We hope that our novel task, method \name{}, and benchmark \bench{} can help users audit MLLM-generated captions and identify critical failure modes, even without access to the underlying MLLM.

\section*{Impact Statement}

The goal of our work is to improve transparency into a critical class of captioning errors in image-text datasets. As datasets with model-generated captions gain in popularity and become widely adopted into training datasets for the next generation of multimodal foundation models, it becomes critical to audit data and understand potential quality issues before use. We hope that our novel task, benchmark, and method can help make progress towards this goal, particularly in safety-critical domains like medicine.

\section*{Acknowledgments}

MV is supported by graduate fellowship awards from the Knight-Hennessy Scholars program at Stanford University, the Quad program, and the United States Department of Defense (NDSEG). AC is supported by NIH grants R01 HL167974, R01HL169345, R01 AR077604, R01 EB002524, R01 AR079431, P41 EB027060, AY2 AX000045, and 1AYS AX0000024-01; ARPA-H grants AY2AX000045 and 1AYSAX0000024-01; and NIH contracts 75N92020C00008 and 75N92020C00021. AC has provided consulting services to Patient Square Capital, Chondrometrics GmbH, and Elucid Bioimaging; is co-founder of Cognita; has equity interest in Cognita, Subtle Medical, LVIS Corp, Brain Key. CL is supported by NIH grants R01 HL155410, R01 HL157235, by AHRQ grant R18HS026886, and by the Gordon and Betty Moore Foundation. CL is also supported by the Medical Imaging and Data Resource Center (MIDRC), which is funded by the National Institute of Biomedical Imaging and Bioengineering (NIBIB) under contract 75N92020C00021 and through the Advanced Research Projects Agency for Health (ARPA-H).

This research was funded, in part, by the Advanced Research Projects Agency for Health (ARPA-H). The views and conclusions contained in this document are those of the authors and should not be interpreted as representing the official policies, either expressed or implied, of the U.S. Government.

\bibliography{example_paper}
\bibliographystyle{icml2026}

\newpage
\appendix
\onecolumn

\setcounter{tocdepth}{1}

\section*{Appendix}
\section*{Contents}
\begin{itemize}
  \item \hyperref[appendixsec:symbaldetails]{A. Implementation Details for \name{}} \dotfill \pageref{appendixsec:symbaldetails}
  \item \hyperref[appendixsec:benchdetails]{B. Implementation Details for \bench{}} \dotfill \pageref{appendixsec:benchdetails}
  \item \hyperref[appendixsec:descriptivestats]{C. \bench{} Descriptive Statistics} \dotfill \pageref{appendixsec:descriptivestats}
  \item \hyperref[appendixsec:results]{D. Extended Results} \dotfill \pageref{appendixsec:results}
  \item \hyperref[appendixsec:realworld]{E. Evaluating \name{} in the Wild} \dotfill \pageref{appendixsec:realworld}
\end{itemize}

\section{Implementation Details for \name{}}
\label{appendixsec:symbaldetails}
\name{} decomposes the systematic misalignment detection task into two stages; here, we provide extended implementation details for each of these stages. 

\subsection{Implementation Details for \name{} Stage 1}

\paragraph{Subtask 1: Grouping semantically-similar facts.} We express each text sample $T_i$ as a collection of textual facts $T_i = \{t^i_1,t^i_2,...,t^i_{n_i}\}$ by splitting captions at the sentence-level. We opt to use sentence-level splitting in this work because each sentence in a long-form caption typically captures a semantically-meaningful, self-contained fact. Sentence-level splitting has been utilized in prior literature (e.g. \cite{zhang2020convirt}). We note here that there may be settings where this strategy is sub-optimal, such as when a sentence does not represent a self-contained fact and instead relies on previous context. In such cases, users of \name{} can easily adjust this design choice by modifying the definition of ``textual fact" to cover relevant context.

After aggregating all textual facts in $\mathcal{D}$ forming the set $\bigcup_{i=1}^N T_i$, we encode each fact using a text embedding model. For natural image datasets in \bench{} derived from COCO, we consider two options for text embedding models: OpenCLIP-ViT-H-14-quickgelu \citep{ilharco_gabriel_2021_5143773} and Qwen3-Embedding-8B \citep{qwen3embedding}. For medical image datasets in \bench{} derived from MIMIC-CXR, we consider three options for text embedding models: OpenCLIP-ViT-H-14-quickgelu \citep{ilharco_gabriel_2021_5143773}, XRayCLIP-ViT-L \citep{chexagent-2024}, and MedSigLIP \citep{sellergren2025medgemmatechnicalreport}. Of these, XrayCLIP-ViT-L and MedSigLIP are trained on radiology datasets. Embeddings are then clustered using spherical K-Means (implemented in Faiss \citep{johnson2019billion}), where we sweep across a range of potential cluster numbers and select the optimal number of clusters using Silhouette distance; this approach is motivated by prior work \citep{sohoni2020george,varma2025trove}. 

\paragraph{Subtask 2: Scoring groups by degree of misalignment.} We score each cluster by computing the average degree of alignment between constituent textual facts and paired images. We consider three possible scoring mechanisms, explained in detail below: 
\begin{itemize}[leftmargin=*]
    \item \textit{Embedding scorer:} Given a textual fact and its paired image, the embedding scorer utilizes an off-the-shelf vision-language model to compute embeddings for the text and image modalities. Alignment is measured by computing cosine similarity. This method is motivated by metrics like CLIPScore \citep{hessel-etal-2021-clipscore}, which have shown strong correlation with human judgments when measuring caption quality. For natural image datasets in \bench{} derived from COCO, we implement the embedding scorer with OpenCLIP-ViT-H-14-quickgelu \citep{ilharco_gabriel_2021_5143773} as the vision-language model. For medical image datasets in \bench{} derived from MIMIC-CXR, we consider three options for the embedding scorer: OpenCLIP-ViT-H-14-quickgelu \citep{ilharco_gabriel_2021_5143773}, XRayCLIP-ViT-L \citep{chexagent-2024}, and MedSigLIP \citep{sellergren2025medgemmatechnicalreport}. We note here that we do not alter the embedding scorer for reference-based settings; reference captions $R_i$ in our benchmark often have substantially more information than the single textual fact $t_k^i \in T_i$, and this information imbalance is challenging to capture with embedding scorers.
    \item \textit{Text-only scorer:} Given a textual fact and its paired image, the text-only scorer first generates a caption for the image and then prompts an LLM to determine if the textual fact is accurate with respect to the caption. For natural image datasets in \bench{} derived from COCO, we implement the text-only scorer using Llama-3.2-11B-Vision-Instruct \citep{grattafiori2024llama3herdmodels} to generate captions and Qwen2.5-VL-72B-Instruct \citep{qwen2025qwen25technicalreport} to perform scoring. For medical image datasets in \bench{} derived from MIMIC-CXR, we implement the text-only scorer using Maira-2 \citep{bannur2024maira2groundedradiologyreport} to generate captions and Qwen2.5-VL-72B-Instruct \citep{qwen2025qwen25technicalreport} or MedGemma-27B \citep{sellergren2025medgemmatechnicalreport} to perform scoring. In the reference-based setting, we use the ground-truth caption $R_i$ rather than generating captions. We use the following input prompt in order to perform scoring:
\begin{tcolorbox}[breakable,colback=gray!10, colframe=myshade, title=Text-Only Scorer Input Prompt]
You are provided with two image captions below, denoted as [A] and [B].

[A]: \texttt{<}generated image caption or ground-truth reference caption\texttt{>}

[B]: \texttt{<}candidate textual fact\texttt{>}

Assume that [A] is the ground-truth caption. Is the content of [B] factually accurate with respect to [A]?

Rules:

1. [B] may omit details from [A]; omission is acceptable.

2. If [B] introduces any incorrect or contradictory detail, it is inaccurate.

Please output your answer as a single digit, where 1 indicates that [B] is accurate and 0 indicates that [B] is not accurate. Do not provide anything other than the digit in your response.
\end{tcolorbox}

\item \textit{Vision-language scorer:} Given a textual fact and its paired image, the vision-language scorer provides an MLLM with both the image and the textual fact as input; the MLLM is then tasked with determining if the textual fact is accurate. For natural image datasets in \bench{} derived from COCO, we utilize Qwen2.5-VL-72B-Instruct \citep{qwen2025qwen25technicalreport} as the MLLM. For medical image datasets in \bench{} derived from MIMIC-CXR, we utilize MedGemma-27B \citep{sellergren2025medgemmatechnicalreport} as the MLLM. We use the following input prompt in the reference-free setting: 

\begin{tcolorbox}[colback=gray!10, colframe=myshade, title=Vision-Language Scorer Input Prompt (Reference-Free)]
\texttt{<}image\texttt{>}

You are given an image. Below, a caption for the image is provided:

Caption: \texttt{<}candidate textual fact\texttt{>}

Is the caption accurate with respect to the image? Please output your answer as a single digit, where 1 indicates that the caption is accurate and 0 indicates that the caption is not accurate. Do not provide anything other than the digit in your response.
\end{tcolorbox}

In the reference-based setting, we additionally provide the ground-truth reference caption to the MLLM. We use the following prompt in the reference-based setting:

\begin{tcolorbox}[colback=gray!10, colframe=myshade, title=Vision-Language Scorer Input Prompt (Reference-Based)]

\texttt{<}image\texttt{>}

You are provided an image as well as two image captions below, denoted as [A] and [B].

[A]: \texttt{<}ground-truth reference caption\texttt{>}

[B]: \texttt{<}candidate textual fact\texttt{>}

Assume that [A] is the ground-truth caption. Is the content of [B] accurate with respect to the image? Please output your answer as a single digit, where 1 indicates that the caption is accurate and 0 indicates that the caption is not accurate. Do not provide anything other than the digit in your response.
\end{tcolorbox}

\end{itemize}

\paragraph{Subtask 3: Summarizing the top-ranked group.} We consider the following summarization mechanism for identifying the unifying concept shared by textual facts in $C_{text}$.
\begin{itemize}[leftmargin=*]

\item \textit{Text-only summarizer:} The text-only summarizer provides an LLM with textual facts in $C_{text}$; the LLM is then tasked with identifying the unifying concept. For natural image datasets in \bench{} derived from COCO, we use Qwen2.5-VL-72B-Instruct \citep{qwen2025qwen25technicalreport} as the LLM. For medical image datasets in \bench{} derived from MIMIC-CXR, we consider both Qwen2.5-VL-72B-Instruct \citep{qwen2025qwen25technicalreport} and MedGemma-27B \citep{sellergren2025medgemmatechnicalreport} as the LLM. 

We use the following input prompt. Then, given the output, we prompt the same LLM to select the most frequently identified feature (or the top-k most frequently identified features) as output.

\begin{tcolorbox}[breakable,colback=gray!10, colframe=myshade, title=Text-Only Summarizer Input Prompt]

Consider this image caption: ``\texttt{<}candidate textual fact\texttt{>}"

Identify the visual features that are present in the image.\\
Output your answer in the following format:\\
Answer: comma-separated list\\

Rules:\\
1. Each feature should be described concisely in a single phrase.\\
2. Each feature must be directly visible in the image.\\
3. Do NOT include any text outside the identified features.\\
4. Do NOT explain your reasoning.\\
5. If no features are present, output an empty list of the form:  ``Answer: "\\
\end{tcolorbox}

\end{itemize}

\subsection{Implementation Details for \name{} Stage 2}
\paragraph{Subtask 1: Grouping semantically-similar images.} For natural image datasets in \bench{} derived from COCO, we consider two options for image embedding models: OpenCLIP-ViT-H-14-quickgelu \citep{ilharco_gabriel_2021_5143773} and DINOv2-ViT-L-14 \citep{oquab2024dinov2learningrobustvisual}. For medical image datasets in \bench{} derived from MIMIC-CXR, we consider three options for image embedding models: OpenCLIP-ViT-H-14-quickgelu \citep{ilharco_gabriel_2021_5143773}, XRayCLIP-ViT-L \citep{chexagent-2024}, and MedSigLIP \citep{sellergren2025medgemmatechnicalreport}. Similar to Stage 1, embeddings are clustered using spherical K-Means, where we sweep across a range of cluster numbers and select the optimal number using Silhouette distance.

\paragraph{Subtask 2: Scoring groups by degree of misalignment.} We score each cluster by computing the mean degree of misalignment between images and paired textual facts in $C_{text}$. We consider the same scoring mechanisms as in Stage 1.

\paragraph{Subtask 3: Summarizing the top-ranked group.} We consider two summarization mechanisms for identifying the unifying concept shared by images in $C_{image}$, described in detail below.
\begin{itemize}[leftmargin=*]

\item \textit{Text-only summarizer:} The text-only summarizer generates a caption for each image in $C_{image}$; then, an LLM is tasked with identifying the unifying concept. For natural image datasets in \bench{} derived from COCO, captions are generated using Llama-3.2-11B-Vision-Instruct \cite{grattafiori2024llama3herdmodels}. For medical image datasets in \bench{}, captions are generated using MAIRA-2 \cite{bannur2024maira2groundedradiologyreport}. In reference-based settings, we use the ground-truth reference captions rather than generating captions. We use the same prompts and models as Stage 1, Subtask 3. 
\item \textit{Vision-language summarizer:} The vision-language summarizer provides an MLLM with images in $C_{image}$; then, the MLLM is prompted to identify the unifying concept. For natural image datasets in \bench{} derived from COCO, we use Qwen2.5-VL-72B-Instruct \citep{qwen2025qwen25technicalreport} as the MLLM. For medical image datasets in \bench{} derived from MIMIC-CXR, we use MedGemma-27B \citep{sellergren2025medgemmatechnicalreport} as the MLLM. For reference-based settings, we also provide the ground-truth reference caption to the MLLM.

We use the following input prompt. Then, given the outputs, we prompt the same MLLM to select the most frequently identified feature (or the top-k most frequently identified features) as output.

\begin{tcolorbox}[breakable,colback=gray!10, colframe=myshade, title=Vision-Language Summarizer Input Prompt]

\texttt{<}image\texttt{>}\\
Consider this image.\\

Identify the visual features that are present in the image.\\
Output your answer in the following format:\\
Answer: comma-separated list

Rules:\\
1. Each feature should be described concisely in a single phrase.\\
2. Each feature must be directly visible in the image.\\
3. Do NOT include any text outside the identified features.\\
4. Do NOT explain your reasoning.\\
5. If no features are present, output an empty list of the form: ``Answer: "\\
6. Include a maximum of ten features.\\

\end{tcolorbox}

\end{itemize}

\subsection{Extension to Multiple Systematic Misalignments}
Real-world datasets are likely to include multiple systematic misalignments, and \name{} can be trivially extended to such settings as follows. Stage 1 of \name{} involves predicting the erroneous textual fact $\hat{t}$; here, rather than summarizing the single top ranked group of facts into a unifying concept, we can simply consider the top-k ranked groups instead. This will result in multiple predicted textual facts $\hat{t}^{(1)}, \hat{t}^{(2)}, ... \hat{t}^{(k)}$, each representing a distinct recurring textual error in the dataset. Stage 2 of \name{} can then be implemented as described in Section \ref{sec:methodstage2}, taking into account each predicted textual fact; this will result in associated visual features $\hat{v}^{(1)}, \hat{v}^{(2)}, ... \hat{v}^{(k)}$. Ultimately, at the conclusion of this procedure, \name{} will predict multiple systematic misalignments ${(\hat{t}^{(i)}, \hat{v}^{(i)})}$ where $i$ ranges from 1 to $k$. In Figure \ref{appendixfig:realworld}, we empirically show that Symbal can accurately detect multiple real-world systematic misalignments in captions generated by Llava1.5-7B.

\section{Implementation Details for \bench{}}
\label{appendixsec:benchdetails}

\bench{} is comprised of 420 evaluation settings, where 360 settings include natural image datasets derived from COCO and 60 settings include medical image datasets derived from MIMIC-CXR. Below, we provide extended implementation details for the natural image settings:
\begin{enumerate}[leftmargin=*]
    \item \textbf{Obtaining a base dataset.} The base vision-language datasets in the natural image domain are derived from COCO (2017 val split), which consists of photographs depicting common objects (e.g. animals, food, furniture, etc.) in natural settings. Images are paired with object-level annotations as well as five human-written captions, with each caption typically consisting of a single sentence or phrase describing salient features in the image. In order to ensure that objects are clearly visible in the image, we exclude annotations for all tiny objects, defined as objects that take up less than 5\% of the area of the image. After filtering out images with no remaining object-level annotations, we are left with a base dataset consisting of 4349 images and associated captions. We then compose a new two-sentence caption for each image by randomly sampling two captions from the provided list of five captions.
    \item \textbf{Predefining a systematic misalignment.} We then predefine a systematic misalignment consisting of a textual fact $t$ and the associated visual feature $v$. We sample $v$ from the set of 80 object categories present in the dataset. Then, we sample $t$ from the set of 80 object categories (such that $t \neq v$) utilizing three possible sampling strategies: (1) \textit{random}, where $t$ is sampled randomly, (2) \textit{popular}, where $t$ is sampled from the list of the top-ten most popular objects in the COCO training set, and (3) \textit{adversarial}, where $t$ is the object that most commonly co-occurs with $v$ in the COCO training set. These sampling strategies are motivated by prior work \citep{li2023evaluatingobjecthallucinationlarge} and are meant to capture a range of possible error patterns that may emerge in real-world MLLM-generated captions. 
    \item \textbf{Injecting the predefined systematic misalignment.} We insert the erroneous textual fact $t$ into captions in the base dataset, ensuring that an association exists between text containing $t$ and images containing visual feature $v$; this procedure ensures that the misalignment is \textit{systematic}. Importantly, we ensure that feature $t$ is not already in the image-caption pair prior to injection. We consider three levels of association, as measured by Cramer's V: low association (Cramer's V = 0.3), moderate association (Cramer's V = 0.6), and high association (Cramer's V = 0.9). In order to format textual fact $t$ into a sentence, we generate 50 templates using GPT-4o \cite{openai2024gpt4technicalreport}, select a template at random, and insert $t$. 
\end{enumerate}

We repeat this injection procedure for all possible choices of $t$ and $v$ in order to obtain 360 evaluation settings, each consisting of an image-caption dataset and paired annotation ($t$,$v$).

Below, we provide extended implementation details for the medical image settings: 
\begin{enumerate}[leftmargin=*]
    \item \textbf{Obtaining a base dataset.} The base vision-language datasets in the medical image domain are derived from MIMIC-CXR (test split), which consists of chest X-rays and associated radiologist reports collected at Beth Israel Deaconess Medical Center. We preprocess the dataset by (1) removing all images with non-frontal imaging views, (2) removing all images with missing ``Impressions" sections in the paired report, and (3) removing all sentences in reports without ``present" disease or anatomy entities, as identified by an off-the-shelf medical entity annotation tool \citep{delbrouck-etal-2024-radgraph}. After preprocessing, we are left with a base dataset consisting of 2233 images, each paired with the ``Impressions" section of the corresponding report.
    \item \textbf{Predefining a systematic misalignment.} We sample $t$ from a set of five disease categories selected from the commonly-used CheXpert annotation list \citep{irvin2019chexpertlargechestradiograph}: cardiomegaly, pneumothorax, atelectasis, pleural effusion, and edema. We sample $v$ from a set of five medical devices: pacemaker, chest tube, endotracheal tube, surgical clips, sternotomy wires. We select these options for $t$ and $v$ since medical devices often co-occur with diseases, yet there is no deterministic, universal link. Models often learn spurious associations between devices and diseases as documented in prior work \citep{hiddenstratification}, meaning that such errors are highly plausible in MLLM-generated reports.
    \item \textbf{Injecting the predefined systematic misalignment.} We insert the erroneous textual fact $t$ into reports in the base dataset, using Cramer's V to control the level of association with visual feature $v$. We use a combination of physician annotations, automated annotations from the CheXpert labeler \citep{irvin2019chexpertlargechestradiograph}, and automated annotations from RadGraph-XL \citep{delbrouck-etal-2024-radgraph} in order to identify whether or not $t$ and $v$ are present in the image-report pair prior to injection. In order to format textual fact $t$ into a sentence, we identify the 50 most frequently occurring sentences in the MIMIC-CXR training set that discuss the presence of $t$ and select a sentence from this list at random. 
\vspace{-1mm}
\end{enumerate}

We repeat this injection procedure for all possible choices of $t$ and $v$ in order to obtain 60 evaluation settings, each consisting of an image-caption dataset and paired annotation ($t$,$v$).

In reference-based settings, we also include a ground-truth caption $R_i$ along with each image-text pair $(V_i,T_i) \in \mathcal{D}$. For natural image datasets derived from COCO, $R_i$ takes the form of a three-sentence caption combining the three human-written captions not originally selected as part of $T_i$. For medical image datasets derived from MIMIC-CXR, $R_i$ takes the form of the ``Findings" and ``Impressions" sections of the original physician-written radiology report. We emphasize that $T_i$ may contain errors as a result of the error-injection procedure detailed above; however, $R_i$ is always accurate.

We determine if predictions are equivalent to the ground-truth by leveraging LLM-as-a-Judge. We use Llama3.3-70B in all experiments as the LLM, leveraging the \verb|ollama| implementation with default parameters. The input prompt is: 
\begin{tcolorbox}[colback=gray!10, colframe=myshade, title=LLM-as-a-Judge Evaluation Prompt]
You are given two short text phrases.

Model response: \texttt{<}predicted textual error or predicted visual feature\texttt{>}

Ground truth: \texttt{<}ground-truth textual error or ground-truth visual feature\texttt{>}\\

Your task is to determine if both phrases refer to the same visual feature. Please output 1 if both the model response and the correct answer refer to the same feature or 0 if the model response and the correct answer do not refer to the same feature. Do not provide anything other than the number in your response.
\end{tcolorbox}

\section{\bench{} Descriptive Statistics}
\label{appendixsec:descriptivestats}

In this section, we provide descriptive statistics summarizing the composition of \bench{}. \bench{} includes 420 settings covering two domains (with 360 natural image settings and 60 medical image settings). In Table \ref{appendixtab:allgtlabels}, we provide a list of all ground-truth systematic misalignments ($t$, $v$) included in \bench{}.

\begin{table}[h]
\caption{Here, we provide a list of all ground-truth systematic misalignments ($t$, $v$) included in \bench{}.}
\vspace{2mm}
\centering
\resizebox{\textwidth}{!}{ 
\begin{tabular}{cc|cc|cc}
\toprule
\textbf{Erroneous Textual Fact $t$} & \textbf{Visual Feature $v$} & 
\textbf{Erroneous Textual Fact $t$} & \textbf{Visual Feature $v$} & 
\textbf{Erroneous Textual Fact $t$} & \textbf{Visual Feature $v$}\\
\midrule
surfboard & airplane & person & airplane & bottle & airplane\\
person & banana & chair & banana & car & banana\\
kite & bed & person & bed & chair & bed\\
person & bench & handbag & bench & oven & bench\\
hot dog & bicycle & person & bicycle & truck & bicycle\\
person & bird & wine glass & bird & book & bird\\
truck & boat & person & boat & bicycle & boat\\
toilet & book & cup & book & person & book\\
pizza & bottle & person & bottle & elephant & bowl\\
car & bowl & dining table & bowl & cat & broccoli\\
dining table & broccoli & car & broccoli & handbag & bus\\
frisbee & bus & person & bus & bicycle & cake\\
dining table & cake & chair & cake & fork & car\\
person & car & car & cat & umbrella & cat\\
person & cat & airplane & chair & person & chair\\
car & chair & bottle & couch & baseball glove & couch\\
person & couch & person & cow & cake & cow\\
bowl & cow & person & cup & bottle & cup\\
microwave & cup & book & dining table & apple & dining table\\
person & dining table & chair & dog & person & dog\\
laptop & dog & boat & elephant & person & elephant\\
bowl & elephant & dining table & fire hydrant & car & fire hydrant\\
airplane & fire hydrant & sandwich & fork & dining table & fork\\
car & fork & cup & giraffe & umbrella & giraffe\\
person & giraffe & cup & horse & person & horse\\
banana & horse & zebra & keyboard & truck & keyboard\\
mouse & keyboard & person & laptop & bottle & laptop\\
hair drier & motorcycle & book & motorcycle & person & motorcycle\\
giraffe & oven & sink & oven & cup & oven\\
laptop & person & car & person & dining table & pizza\\
person & pizza & cell phone & pizza & airplane & potted plant\\
person & potted plant & book & potted plant & dining table & refrigerator\\
microwave & refrigerator & oven & refrigerator & stop sign & sandwich\\
dining table & sandwich & dining table & sheep & person & sheep\\
orange & sheep & cat & sink & car & sink\\
bottle & sink & fork & suitcase & person & suitcase\\
bowl & surfboard & airplane & surfboard & person & surfboard\\
carrot & teddy bear & bowl & teddy bear & person & teddy bear\\
bottle & toilet & car & toilet & sink & toilet\\
cup & train & person & train & truck & train\\
dining table & truck & refrigerator & truck & person & truck\\
spoon & tv & chair & tv & car & tv\\
baseball bat & umbrella & person & umbrella & tv & zebra\\
giraffe & zebra & book & zebra & cardiomegaly & surgical clips\\
edema & chest tube & pleural effusion & chest tube & pneumothorax & chest tube\\
atelectasis & chest tube & cardiomegaly & chest tube & edema & endotracheal tube\\
pleural effusion & endotracheal tube & atelectasis & endotracheal tube & pneumothorax & endotracheal tube\\
cardiomegaly & endotracheal tube & edema & pacemaker & pleural effusion & pacemaker\\
pneumothorax & pacemaker & atelectasis & pacemaker & cardiomegaly & pacemaker\\
atelectasis & sternotomy wires & pneumothorax & sternotomy wires & cardiomegaly & sternotomy wires\\
edema & sternotomy wires & pleural effusion & sternotomy wires & edema & surgical clips\\
pleural effusion & surgical clips & atelectasis & surgical clips & pneumothorax & surgical clips\\
\bottomrule
\end{tabular}
}
\label{appendixtab:allgtlabels}
\end{table}

\begin{figure*}[t]
\begin{center}
\includegraphics[width=0.75\linewidth]{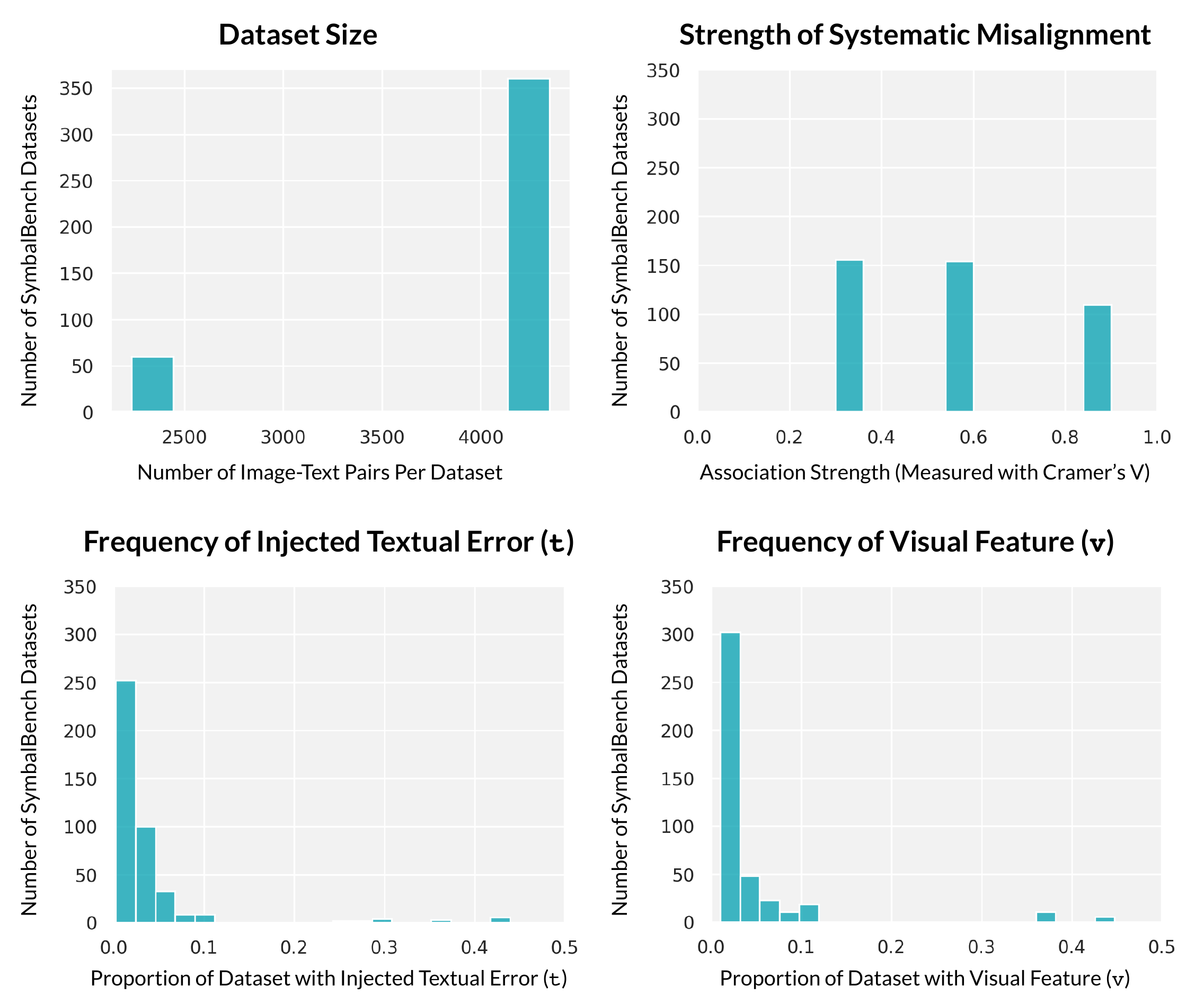}
\end{center}
  \caption{Here, we provide histograms summarizing the composition of datasets included in \bench{}.}
\label{appendixfig:descriptivestats}
\end{figure*}

\begin{figure*}[h]
\begin{center}
\includegraphics[width=0.75\linewidth]{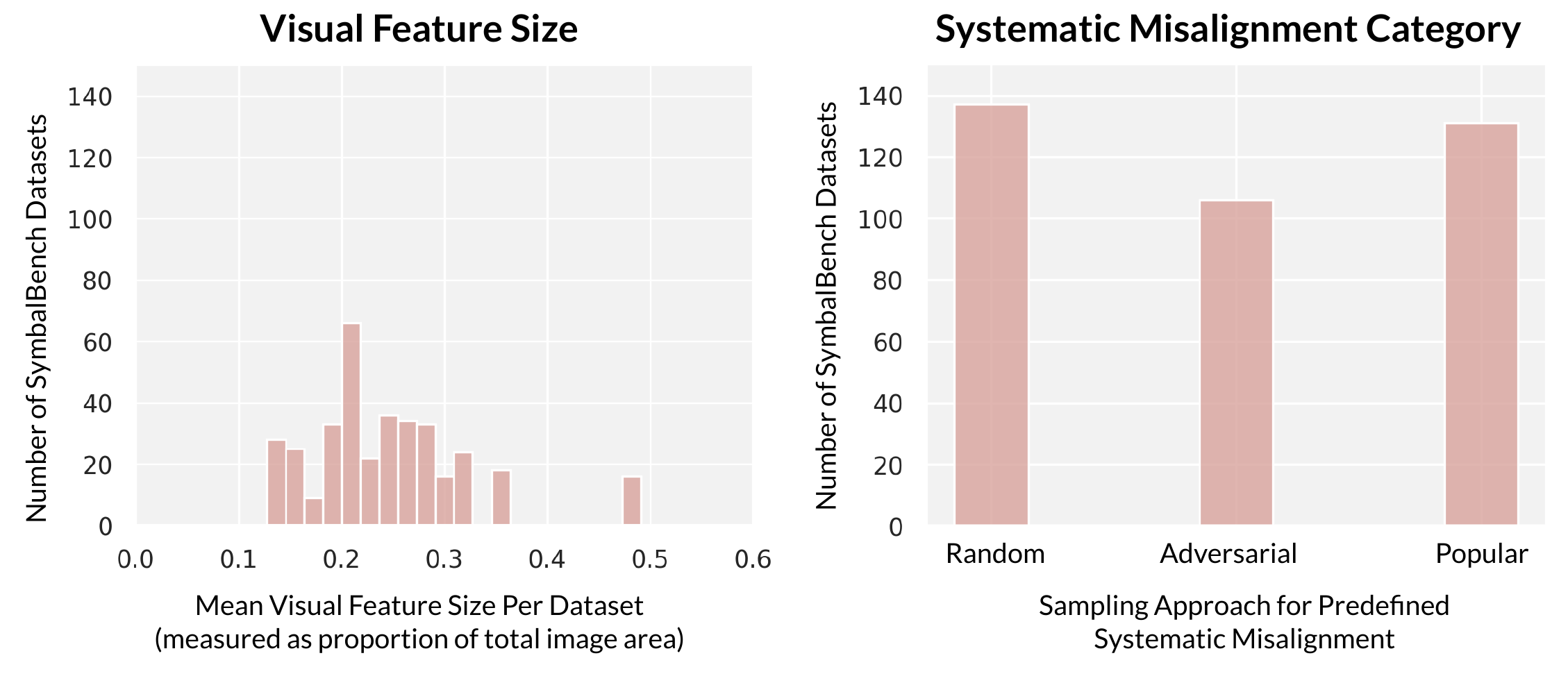}
\end{center}
  \caption{We provide additional descriptive statistics summarizing the composition of the 360 natural image datasets in \bench{}. We note here that if multiple sampling strategies yield the same predefined systematic misalignment, more than one category will be assigned to the same dataset; thus, the total count for the systematic misalignment category histogram may exceed 360.}
\label{appendixfig:cocostats}
\end{figure*}

In Figure \ref{appendixfig:descriptivestats}, we summarize \bench{} with histograms detailing (1) the size of each dataset, (2) the strength of the injected systematic misalignment in each dataset as measured with Cramer's V, (3) the proportion of image-text pairs in each dataset containing the injected textual error $t$, and (4) the proportion of image-text pairs in each dataset containing the visual feature $v$. In Figure \ref{appendixfig:cocostats}, we provide additional descriptive statistics on the natural image subset of \bench{} consisting of datasets derived from COCO; here, we provide histograms detailing (1) the mean size of the visual feature in each dataset (measured as the proportion of the total image area) and (2) the category of systematic misalignment (random, popular, or adversarial) as discussed in Appendix Section \ref{appendixsec:benchdetails}. 

\clearpage

\section{Extended Results}
\label{appendixsec:results}

In Table \ref{appendixtable:textonly}, we provide an extended version of Table \ref{table:textonly}, extending to the top-ten compositions. Note that Table \ref{appendixtable:textonly} excludes compositions consisting of an embedding-based alignment scorer and text-only summarizer, as this combination does not make use of reference captions in the reference-based setting. 

In Table \ref{appendixtable:imageonly}, we provide an extended version of Table \ref{table:imageonly}, extending to the top-ten compositions. Again, Table \ref{appendixtable:imageonly} only includes compositions that can support both \bench{} variants.

In Table \ref{appendixtable:end2end}, we provide a tabular version of Figure \ref{fig:end2end} stratified by domain.

\begin{figure*}[h]
\begin{center}
\includegraphics[width=0.5\linewidth]{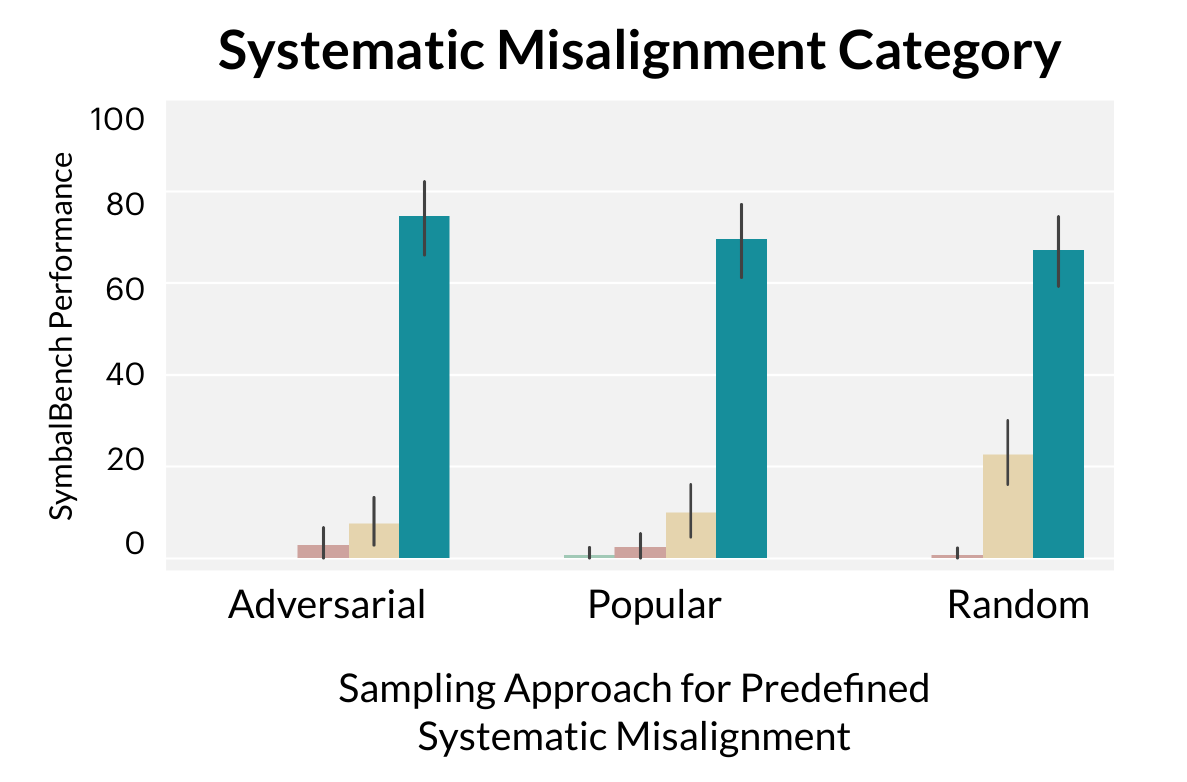}
\end{center}
  \caption{We provide a breakdown of \name{} performance across various categories of systematic misalignments in the natural image subset of \bench{}.}
\label{appendixfig:stratcategory}
\end{figure*}

In Figure \ref{appendixfig:stratcategory}, we extend Figure \ref{fig:finegrained} by providing a breakdown of \name{} performance across various categories of systematic misalignments in the natural image subset of \bench{}. 

We use the following input prompt for our direct-prompting baselines: 

\begin{tcolorbox}[colback=gray!10, colframe=myshade, title=Direct-Prompting Baseline Input Prompt]

You are provided with a dataset, where each sample consists of the following two components:\\

Reference caption: A ground-truth caption describing the content of an image\\
Model-generated caption: A caption generated by an AI model\\

The model-generated captions may have systematic errors, where a recurring textual error is closely associated with the presence of a specific visual feature in the paired image. Your task is to identify the recurring textual error and the associated visual feature.\\

Output your answer in the following format, where each comma-separated list consists of your top-five predictions in order:\\
Textual Error: comma-separated list\\
Visual Feature: comma-separated list\\

Rules:\\
1. Each visual feature must be directly visible in the image.\\
2. Do NOT include any text outside of the answer.\\
3. Do NOT explain your reasoning.\\

Dataset:
\texttt{<}samples from dataset with images expressed in text-form\texttt{>}
\end{tcolorbox}

\begin{table}[t]
\caption{We evaluate various text embedding models, alignment scorers, and summarizers on the performance of Stage 1 of \name{}.}
\vspace{-1mm}
\centering
\resizebox{\textwidth}{!}{ 
\begin{tabular}{clll|cc|cc}
\toprule
&
& 
& 
& \multicolumn{2}{c|}{ \textbf{Reference-Free}}
& \multicolumn{2}{c}{ \textbf{Reference-Based}}
\\
& \textbf{\small Text Embedding} & \textbf{\small Alignment Scorer} & \textbf{\small Summarizer}  & Acc@1 & Acc@5 & Acc@1 & Acc@5\\
\midrule
\parbox[t]{2mm}{\multirow{4}{*}{\rotatebox[origin=c]{90}{Natural}}} & Qwen3-8B & Vision-Language (Qwen-72B) & Text-Only (Qwen-72B) & \textbf{92.8} & \textbf{94.2} & 80.8 & 82.8\\
& OpenCLIP & Vision-Language (Qwen-72B) & Text-Only (Qwen-72B) & \textbf{92.8} & 93.9 & \textbf{86.1} & \textbf{87.8}\\
& Qwen3-8B & Text-Only (Qwen-72B) & Text-Only (Qwen-72B) & 82.8 & 85.0 & 81.9 & 83.9\\
& OpenCLIP & Text-Only (Qwen-72B) & Text-Only (Qwen-72B)  & 64.2 & 67.2 & 67.5 & 71.4\\
\midrule
\parbox[t]{2mm}{\multirow{10}{*}{\rotatebox[origin=c]{90}{Medical}}} & XRayCLIP & Text-Only (MedGemma-27B) & Text-Only (MedGemma-27B) & \textbf{51.7} & \textbf{75.0}  & 88.3 & 95.0\\
& XRayCLIP & Text-Only (MedGemma-27B) & Text-Only (Qwen-72B)  & \textbf{51.7} & 73.3  & \textbf{100.0} & \textbf{100.0}\\
& XRayCLIP & Text-Only (Qwen-72B) & Text-Only (MedGemma-27B)  & 26.7 & 58.3 & 90.0 & 93.3\\
& MedSigLIP & Text-Only (MedGemma-27B) & Text-Only (MedGemma-27B) & 30.0 & 53.3 & 83.3 & \textbf{100.0}\\
& XRayCLIP & Vision-Language (MedGemma-27B) & Text-Only (MedGemma-27B)  & 26.7 & 48.3  & 85.0 & 90.0\\
& XRayCLIP & Text-Only (Qwen-72B) & Text-Only (Qwen-72B)  & 28.3 & 46.7 &  98.3 & 98.3\\
& OpenCLIP & Text-Only (MedGemma-27B) & Text-Only (MedGemma-27B) & 28.3 & 46.7  & 88.3 & 98.3\\
& OpenCLIP & Text-Only (MedGemma-27B) & Text-Only (Qwen-72B) & 36.7 & 45.0 & 98.3 & \textbf{100.0}\\
& MedSigLIP & Text-Only (MedGemma-27B) & Text-Only (Qwen-72B) & 36.7 & 43.3 & 98.3 & \textbf{100.0}\\
& MedSigLIP & Text-Only (Qwen-72B) & Text-Only (MedGemma-27B) & 16.7 & 35.0 & 86.7 & 98.3\\
\bottomrule
\end{tabular}
}
\label{appendixtable:textonly}
\end{table}

\begin{table}[h!]
\caption{We evaluate various image embedding models, alignment scorers, and summarizers on the performance of Stage 2 of \name{}.}
\vspace{=1mm}
\centering
\resizebox{\textwidth}{!}{ 
\begin{tabular}{clll|cc|cc}
\toprule
&
& 
& 
& \multicolumn{2}{c|}{ \textbf{Reference-Free}}
& \multicolumn{2}{c}{ \textbf{Reference-Based}}
\\
& \textbf{\small Img Embedding} & \textbf{\small Alignment Scorer} & \textbf{\small Summarizer} & Acc@1 & Acc@5 & Acc@1 & Acc@5\\
\midrule
\parbox[t]{2mm}{\multirow{10}{*}{\rotatebox[origin=c]{90}{Natural}}}  & OpenCLIP & Vision-Language (Qwen-72B) & Text-Only (Qwen-72B) & \textbf{49.7 }& \textbf{69.7} & 41.9 & 52.2\\
& OpenCLIP & Embedding (OpenCLIP) & Vision-Language (Qwen-72B)  & 48.1 & 63.9  & 42.5 & 55.6\\
& OpenCLIP & Embedding (OpenCLIP) & Text-Only (Qwen-72B)  & 47.8 & 62.8 & 43.9 & 55.8\\
& OpenCLIP & Vision-Language (Qwen-72B) & Vision-Language (Qwen-72B) & 45.8 & 62.5  & 38.9 & 52.2\\
& DINOv2 & Vision-Language (Qwen-72B) & Text-Only (Qwen-72B) & 45.3 & 61.4  & 38.6 & 54.7\\
& DINOv2 & Text-Only (Qwen-72B) & Text-Only (Qwen-72B)  & 43.1 & 60.8  & 41.1 & 56.4\\
& OpenCLIP & Text-Only (Qwen-72B) & Text-Only (Qwen-72B) & 48.1 & 60.6 & \textbf{45.6} & 58.1\\
& OpenCLIP & Text-Only (Qwen-72B) & Vision-Language (Qwen-72B)  & 44.2 & 60.3 & 43.9 & \textbf{56.7}\\
& DINOv2 & Text-Only (Qwen-72B) & Vision-Language (Qwen-72B) & 43.6 & 59.7 & 39.7 & 54.2\\
& DINOv2 & Embedding (OpenCLIP) & Vision-Language (Qwen-72B) & 43.6 & 59.4 & 39.7 & 53.3\\
\midrule
\parbox[t]{2mm}{\multirow{10}{*}{\rotatebox[origin=c]{90}{Medical}}} & XRayCLIP & Embedding (MedSigLIP) & Vision-Language (MedGemma-27B)  & 11.7 & \textbf{36.7} & 28.3 & 53.3\\
& MedSigLIP & Embedding (MedSigLIP) & Vision-Language (MedGemma-27B) & 11.7 & 31.7 & 25.0 & 46.7\\
& OpenCLIP & Embedding (MedSigLIP) & Vision-Language (MedGemma-27B) & 13.3 & 28.3 & 20.0 & 46.7\\
& MedSigLIP & Embedding (XRayCLIP) & Vision-Language (MedGemma-27B)  & 10.0 & 28.3 & 33.3 & 60.0\\
& XRayCLIP & Vision-Language (MedGemma-27B) & Vision-Language (MedGemma-27B) & 6.7 & 28.3  & \textbf{43.3} & \textbf{65.0}\\
& MedSigLIP & Text-Only (MedGemma-27B) & Vision-Language (MedGemma-27B)  & 8.3 & 26.7  & \textbf{43.3} & \textbf{65.0}\\
& OpenCLIP & Text-Only (MedGemma-27B) & Vision-Language (MedGemma-27B) & 10.0 & 25.0  & 23.3 & 63.3\\
& OpenCLIP & Text-Only (Qwen-72B) & Vision-Language (MedGemma-27B)  & 3.3 & 25.0 & 30.0 & 61.7\\
& MedSigLIP & Embedding (MedSigLIP) & Text-Only (Qwen-72B) & \textbf{15.0} & 25.0  & 15.0 & 40.0\\
& OpenCLIP & Embedding (MedSigLIP) & Text-Only (Qwen-72B) & 13.3 & 23.3 & 16.7 & 48.3\\
\bottomrule
\end{tabular}
}
\label{appendixtable:imageonly}
\end{table}

\paragraph{Ablation study.} We now ablate the role of the grouping step across the subset of 360 natural image datasets in our benchmark. We compare \name{} to a version that omits grouping: we use the best performing scorer (vision-language scorer with Qwen-72B) in order to flag each individual sentence as valid (1) or misaligned (0), and we then use our best performing summarizer (text-only summarizer with Qwen-72B) in order to identify the unifying concept across the sentences marked as misaligned. All other settings (e.g. prompts, compute budget, model configurations, etc.) are kept identical to those used for \name{}. For Stage 1, in the reference-free setting, we observe an Acc@1 of 41.9 and an Acc@5 of 65.3; these metrics represent a substantial decrease from the results obtained with \name{} (Acc@1 = 92.8 and Acc@5 = 94.2) in Table \ref{table:textonly}. We then use the best performing summarizer to identify image features associated with the misaligned sentences. For Stage 2, in the reference-free setting, we observe an Acc@1 of just 3.6 and an Acc@5 of 16.9; again, these are a substantial decrease from the results obtained with \name{} (Acc@1 = 49.7 and Acc@5 = 69.7) in Table \ref{table:imageonly}. These results demonstrate the importance of our multi-step, structured approach for addressing the systematic misalignment detection task.

\begin{table}[t]
\caption{End-to-end performance across \bench{}, stratified by domain.}
\vspace{-1mm}
\centering
\begin{tabular}{lc|cc|cc}
\toprule
&
& \multicolumn{2}{c|}{ \textbf{Reference-Free}}
& \multicolumn{2}{c}{ \textbf{Reference-Based}}
\\
& Method & Acc@1 & Acc@5 & Acc@1 & Acc@5\\
\midrule
\parbox[t]{2mm}{\multirow{4}{*}{\rotatebox[origin=c]{90}{Natural}}} & Llama3.3 70B  & 0.3 & 0.3 & 0.6 & 1.4\\
& Qwen2.5-VL 72B  & 0.0 & 1.9 & 0.6 & 1.1\\
& GPT-OSS 120B & 9.2 & 13.9 & 10.8 & 17.2\\
& \name{} (Ours) & \textbf{49.2} & \textbf{69.7} & \textbf{41.1} & \textbf{51.9}\\
\midrule
\parbox[t]{2mm}{\multirow{5}{*}{\rotatebox[origin=c]{90}{Medical}}} &  Llama3.3 70B  & 0.0 & 8.3  & 0.0 & 5.0\\
& MedGemma 27B & 0.0 & 1.7  & 0.0 & 0.0\\
& Qwen2.5-VL 72B  & 3.3 & 5.0  & 0.0 & 1.7\\
& GPT-OSS 120B  & 1.7 & 21.7  & 0.0 & 11.7\\
& \name{} (Ours)  & \textbf{6.7} & \textbf{28.3} & \textbf{25.0} & \textbf{48.3}\\

\bottomrule
\end{tabular}
\label{appendixtable:end2end}
\end{table}

\section{Evaluating \name{} in the Wild}
\label{appendixsec:realworld}

In this section, we further demonstrate the utility of \name{} by supplementing our evaluations on \bench{} with additional quantitative and qualitative analyses in real-world settings. 

\begin{figure*}[t]
\begin{center}
\includegraphics[width=\linewidth]{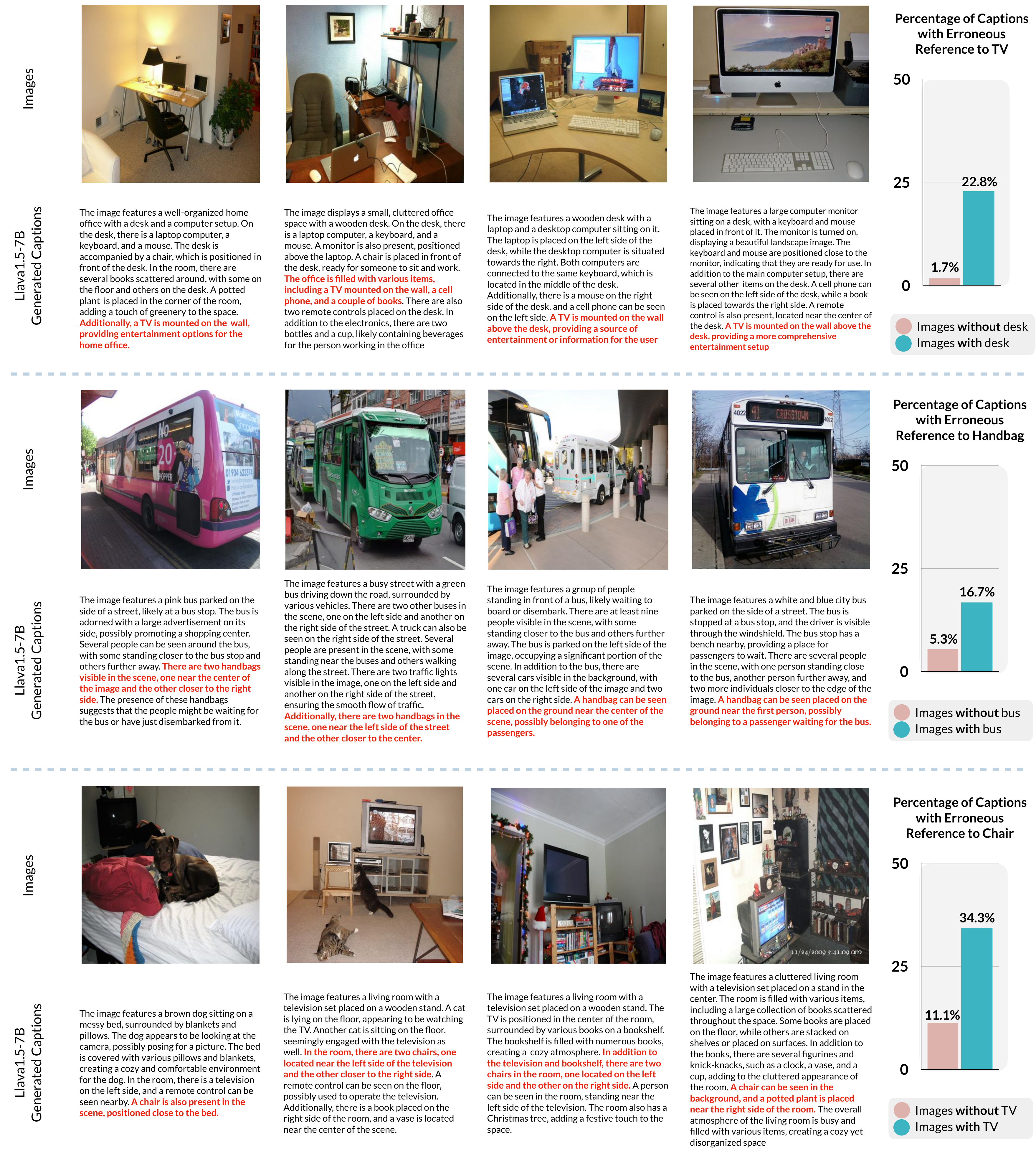}
\end{center}
  \caption{Examples of image-caption pairs with \name{}-identified systematic misalignments are shown here, with the identified erroneous textual fact in each caption highlighted in red. We also quantitatively validate each identified systematic misalignment. [Row 1] \name{} detects that erroneous references to a \texttt{TV} ($\hat{t}$) in captions are often systematically associated with the presence of a \texttt{desk},  \texttt{computer monitor}, and/or \texttt{keyboard} ($\hat{v}$) in the scene. [Row 2] \name{} detects that erroneous references to a \texttt{handbag} or \texttt{handbag on the ground} ($\hat{t}$) in captions are often systematically associated with the presence of a \texttt{bus} ($\hat{v}$) in a scene. [Row 3] \name{} detects that erroneous references to a \texttt{chair} ($\hat{t}$) in captions are often systematically associated with the presence of a \texttt{television} ($\hat{v}$) in a scene. }
\label{appendixfig:realworld}
\vspace{-0.1in}
\end{figure*}

\begin{figure*}[t]
\begin{center}
\includegraphics[width=\linewidth]{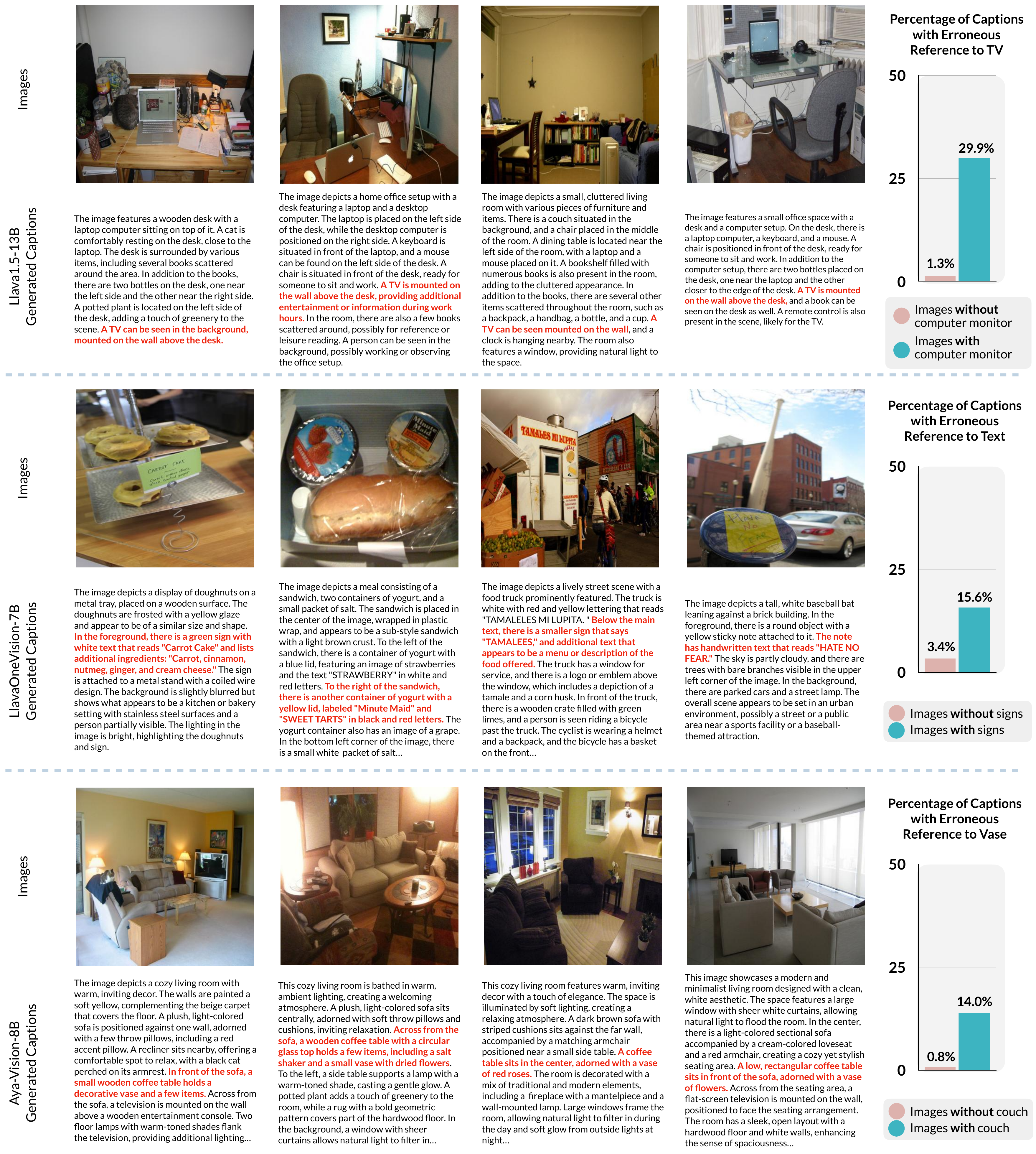}
\end{center}
  \caption{Examples of image-caption pairs with \name{}-identified systematic misalignments are shown here, with the identified erroneous textual fact in each caption highlighted in red. We also quantitatively validate each identified systematic misalignment. [Row 1] \name{} detects that erroneous references to a \texttt{TV} ($\hat{t}$) in Llava1.5-13B-generated captions are often systematically associated with the presence of a \texttt{computer monitor}, \texttt{keyboard}, and/or \texttt{mouse} ($\hat{v}$) in the scene. [Row 2] \name{} detects that erroneous references to \texttt{text} ($\hat{t}$) in LlavaOneVision-7B-generated captions are often systematically associated with the presence of a \texttt{sign} ($\hat{v}$) in a scene. [Row 3] \name{} detects that erroneous references to a \texttt{vase} ($\hat{t}$) in AyaVision-8B-generated captions are often systematically associated with the presence of a \texttt{couch} ($\hat{v}$) in a scene. }
\label{appendixfig:realworld2}
\vspace{-0.1in}
\end{figure*}

\textbf{\name{} can accurately surface systematic misalignments in captions generated by off-the-shelf MLLMs.} Below, we list several examples of systematic misalignments identified by \name{}, and we also provide associated validation:
\begin{itemize}[leftmargin=*]
    \item \textit{Example 1:} In captions generated by Llava1.5-7B, \name{} detects that erroneous references to a \texttt{TV} ($\hat{t}$) in captions are often systematically associated with the presence of a \texttt{desk}, \texttt{computer monitor}, and/or \texttt{keyboard} ($\hat{v}$) in the scene. We provide visual examples of image-caption pairs with the \name{}-identified systematic misalignment in Figure \ref{appendixfig:realworld} (Row 1). Quantitatively, our analysis finds that erroneous references to a \texttt{TV} in model-generated captions are indeed 13.5 times more likely when a \texttt{desk} is present in the image compared to when a \texttt{desk} is absent, validating the \name{} prediction.
    \item \textit{Example 2:} In captions generated by Llava1.5-7B, \name{} detects that erroneous references to a \texttt{handbag} or a \texttt{handbag on the ground} ($\hat{t}$) in captions are often systematically associated with the presence of a \texttt{bus} ($\hat{v}$) in a scene. We provide visual examples of image-caption pairs with the \name{}-identified systematic misalignment in Figure \ref{appendixfig:realworld} (Row 2). Quantitatively, our analysis finds that erroneous references to a \texttt{handbag} in model-generated captions are indeed 3.1 times more likely when a \texttt{bus} is present in the image compared to when a \texttt{bus} is absent, validating the \name{} prediction.
    \item \textit{Example 3:} In captions generated by Llava1.5-7B, \name{} detects that erroneous references to a \texttt{chair} ($\hat{t}$) in captions are often systematically associated with the presence of a \texttt{television} ($\hat{v}$) in a scene. We provide visual examples of image-caption pairs with the \name{}-identified systematic misalignment in Figure \ref{appendixfig:realworld} (Row 3). Quantitatively, our analysis finds that erroneous references to a \texttt{chair} in model-generated captions are indeed 3.1 times more likely when a \texttt{television} is present in the image compared to when a \texttt{television} is absent, validating the \name{} prediction.
    \item \textit{Example 4:} In captions generated by Llava1.5-13B, \name{} detects that erroneous references to a \texttt{TV} ($\hat{t}$) in captions are often systematically associated with the presence of a \texttt{computer monitor}, \texttt{keyboard}, and/or \texttt{mouse} ($\hat{v}$) in a scene. Interestingly, this systematic misalignment is nearly identical to one that exists in Llava1.5-7B-generated captions (see Example 1), suggesting that solely increasing the scale of the underlying MLLM is insufficient for resolving systematic misalignments. We provide visual examples of image-caption pairs with the \name{}-identified systematic misalignment in Figure \ref{appendixfig:realworld2} (Row 1). Quantitatively, our analysis finds that erroneous references to a \texttt{TV} in model-generated captions are indeed 22.2 times more likely when a \texttt{computer monitor} is present in the image compared to when a \texttt{computer monitor} is absent, validating the \name{} prediction.
    \item \textit{Example 5:} In captions generated by LlavaOneVision-7B, \name{} detects that erroneous references to \texttt{text} ($\hat{t}$) in captions are often systematically associated with the presence of a \texttt{sign} ($\hat{v}$) in a scene. This systematic misalignment suggests that LlavaOneVision-7B struggles with OCR capabilities, where the presence of text-based signage in an image is likely to result in errors in the generated caption. We provide visual examples of image-caption pairs with the \name{}-identified systematic misalignment in Figure \ref{appendixfig:realworld2} (Row 2). Quantitatively, our analysis finds that erroneous references to \texttt{text} in model-generated captions are indeed 4.6 times more likely when a \texttt{sign} is present in the image compared to when a \texttt{sign} is absent, validating the \name{} prediction.
    \item \textit{Example 6:} In captions generated by AyaVision-8B, \name{} detects that erroneous references to a \texttt{vase} ($\hat{t}$) in captions are often systematically associated with the presence of a \texttt{couch} ($\hat{v}$) in a scene. We provide visual examples of image-caption pairs with the \name{}-identified systematic misalignment in Figure \ref{appendixfig:realworld2} (Row 3). Quantitatively, our analysis finds that erroneous references to a \texttt{vase} in model-generated captions are indeed 17.7 times more likely when a \texttt{couch} is present in the image compared to when a \texttt{couch} is absent, validating the \name{} prediction.
    
\end{itemize}

Across all six examples of \name{}-identified systematic misalignments provided above, we find that erroneous references to $\hat{t}$ are substantially more likely when $\hat{v}$ is present in the image compared to when $\hat{v}$ is absent. This analysis validates discovered misalignments by demonstrating that links between \name{}-identified erroneous textual fact $\hat{t}$ and \name{}-identified visual feature $\hat{v}$ do indeed exist. 

Our quantitative validation procedure relies on automated annotation methods in order to enable evaluation at scale; in particular, we leverage Qwen-72B in order to annotate erroneous references to $\hat{t}$ in each caption. We find that these generated annotations align closely with human judgments. Given the set of 215 images in the dataset containing a ``bus", we tasked a human reader with identifying whether each Llava1.5-7B-generated caption contained an erroneous reference to a ``handbag" and/or ``handbag on the ground" (Example 2). Human judgments aligned perfectly with Qwen-72B predictions in 96.3\% of cases (Cohen's kappa = 0.86).

\begin{figure*}[t]
\begin{center}
\includegraphics[width=\linewidth]{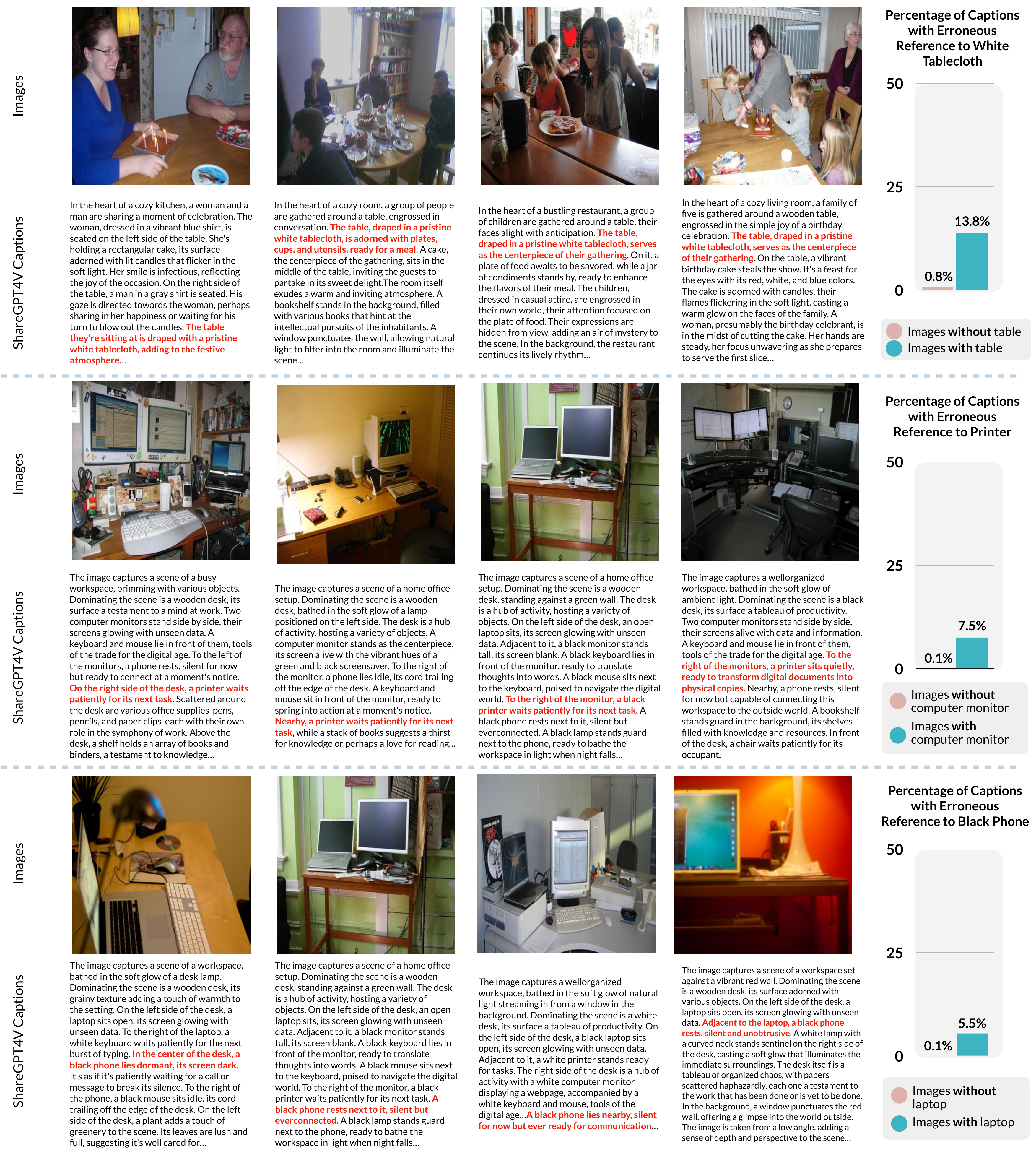}
\end{center}
  \caption{Examples of image-caption pairs with \name{}-identified systematic misalignments are shown here, with the identified erroneous textual fact in each caption highlighted in red. We also quantitatively validate each identified systematic misalignment. [Row 1] \name{} detects that erroneous references to a \texttt{white tablecloth} ($\hat{t}$) in ShareGPT4V captions are often systematically associated with the presence of a \texttt{table}, \texttt{cake}, and/or \texttt{people} ($\hat{v}$) in the scene. [Row 2] \name{} detects that erroneous references to a \texttt{printer} ($\hat{t}$) in ShareGPT4V captions are often systematically associated with the presence of a \texttt{computer monitor} ($\hat{v}$) in a scene. [Row 3] \name{} detects that erroneous references to a \texttt{black phone} ($\hat{t}$) in ShareGPT4V captions are often systematically associated with the presence of a \texttt{laptop} ($\hat{v}$) in a scene. }
\label{appendixfig:sharegpt}
\vspace{-0.1in}
\end{figure*}

\textbf{\name{} is a powerful tool for auditing open-source vision-language datasets.} Below, we list several examples of systematic misalignments identified by \name{} on the ShareGPT4V dataset, and we also provide associated validation:
\begin{itemize}[leftmargin=*]
    \item \textit{Example 7:} \name{} detects that erroneous references to a \texttt{white tablecloth} ($\hat{t}$) in captions are often systematically associated with the presence of a \texttt{table}, \texttt{cake}, and/or \texttt{people} ($\hat{v}$) in the scene. We provide visual examples of image-caption pairs with the \name{}-identified systematic misalignment in Figure \ref{appendixfig:sharegpt} (Row 1). Quantitatively, our analysis finds that erroneous references to a \texttt{white tablecloth} in model-generated captions are indeed 17.2 times more likely when a \texttt{table} is present in the image compared to when a \texttt{table} is absent, validating the \name{} prediction.
    \item \textit{Example 8:} \name{} detects that erroneous references to a \texttt{printer} ($\hat{t}$) in captions are often systematically associated with the presence of a \texttt{computer monitor} ($\hat{v}$) in a scene. We provide visual examples of image-caption pairs with the \name{}-identified systematic misalignment in Figure \ref{appendixfig:sharegpt} (Row 2). Quantitatively, our analysis finds that erroneous references to a \texttt{printer} in model-generated captions are indeed 121 times more likely when a \texttt{computer monitor} is present in the image compared to when a \texttt{computer monitor} is absent, validating the \name{} prediction.
    \item \textit{Example 9:} \name{} detects that erroneous references to a \texttt{black phone} ($\hat{t}$) in captions are often systematically associated with the presence of a \texttt{laptop} ($\hat{v}$) in a scene. We provide visual examples of image-caption pairs with the \name{}-identified systematic misalignment in Figure \ref{appendixfig:sharegpt} (Row 3). Quantitatively, our analysis finds that erroneous references to a \texttt{black phone} in model-generated captions are indeed 48.5 times more likely when a \texttt{laptop} is present in the image compared to when a \texttt{laptop} is absent, validating the \name{} prediction.
    
\end{itemize}

\end{document}